\documentclass[10pt,journal,cspaper,compsoc]{IEEEtran}

\ifCLASSOPTIONcompsoc
\usepackage[nocompress]{cite}
\else
\usepackage{cite}
\fi

\usepackage{epsfig}
\usepackage{subfigure}
\usepackage{amsmath}
\usepackage{amssymb}
\usepackage{cases}
\usepackage{hhline}
\usepackage{caption}
\usepackage{algorithm}
\usepackage{algorithmic}
\usepackage[T1]{fontenc}
\usepackage{color}
\usepackage{tabularx}
\usepackage{booktabs}
\usepackage{makecell}
\usepackage{multirow}
\usepackage{float}
\usepackage{graphicx}
\usepackage{url}
\usepackage{graphicx}
\usepackage{placeins}
\usepackage{xcolor}
\usepackage{colortbl}

\begin{document}
\title{Classification-Head Bias in Class-Level Machine Unlearning: Diagnosis, Mitigation, and Evaluation}

\author{Weidong Zheng, Kongyang Chen, Yuanwei Guo, Yatie Xiao
\IEEEcompsocitemizethanks{
\IEEEcompsocthanksitem{Weidong Zheng and Yatie Xiao are with School of Computer Science and Cyber Engineering, Guangzhou University, Guangzhou 510335, China.} 
\IEEEcompsocthanksitem{Kongyang Chen is with School of Artificial Intelligence, Guangzhou University, Guangzhou 510006, China. He is also with Pazhou Lab, Guangzhou 510006, China.}
\IEEEcompsocthanksitem{Yuanwei Guo is with Guangzhou Institute of Internet of Things, Guangzhou 511462, China.} 
}}

\IEEEtitleabstractindextext{
\begin{abstract}
Class-level machine unlearning aims to remove the influence of specified classes while preserving model utility on retained classes. Existing methods are commonly evaluated by retain-set accuracy, forget-set accuracy, and unlearning time, but these metrics provide limited insight into how forgetting is achieved internally. In this paper, we reveal a bias-dominated shortcut in class-level unlearning: the prediction of forgotten classes can be suppressed by decreasing the corresponding bias terms in the final classification head. We first analyze the gradient dynamics of classification-head biases under softmax cross-entropy training, explaining why retain-set-only optimization tends to reduce the biases of absent classes. Based on this observation, we introduce BiasShift as a diagnostic baseline, showing that simple bias manipulation can satisfy conventional unlearning metrics while leaving abnormal bias patterns that reveal forgotten labels. To mitigate excessive forgotten-class bias suppression, we propose two bias-aware mechanisms, namely Two-Stage Bias Gradient Reversal Mechanism (TS-BGRM) and Lower-Bound Hinge Regularization (LB-HR). We further introduce three bias-oriented metrics, including Bias Stability Coefficient (BSC), Median Bias Gap (MBG), and Minimal Bias Score (MBS), to quantify bias dependence and potential leakage. Experiments on CIFAR-10, CIFAR-100, and Tiny-ImageNet demonstrate that the proposed methods maintain competitive unlearning performance while producing more stable bias distributions. We have released our code at {https://github.com/zwd2024/Beyond-the-Shadow-of-Bias-From-Classification-Head-Bias-to-Parameter- Redistribution}.
\end{abstract}
}

\maketitle
\IEEEdisplaynontitleabstractindextext
\IEEEpeerreviewmaketitle

\section{Introduction}

With the increasing deployment of machine learning models in real-world applications, training data may contain outdated, low-quality, biased, or privacy-sensitive information. Once such data are found to be inappropriate for continued use, simply keeping their influence inside a trained model may violate privacy requirements, degrade model reliability, or conflict with regulatory obligations. Recent privacy regulations, such as the General Data Protection Regulation (GDPR), the California Consumer Privacy Act (CCPA), and the Data Security Law of the People's Republic of China, have strengthened the requirement that data owners should have the right to request the removal of their data from data-driven systems~\cite{GDPR2016,CCPA2018,DataSecurityLaw2021}. A straightforward solution is to remove the target data from the training set and retrain the model from scratch on the remaining data. However, full retraining is usually computationally expensive and time-consuming, especially when only a small portion of the training data needs to be forgotten.

Machine unlearning has therefore emerged as an important research direction for efficiently removing the influence of specified training data from a trained model\cite{hatami2026class, panda2025partially, gao2026illusion}. Depending on the forgetting granularity, existing studies can be broadly divided into sample-level unlearning and class-level unlearning. In sample-level unlearning, the goal is to remove the influence of individual samples, whereas in class-level unlearning, all samples belonging to one or more specified classes are required to be forgotten~\cite{hatami2026class, qin2026machine,  tarun2023fast}. This paper focuses on class-level unlearning, where the model is expected to lose its predictive capability on the forgotten classes while maintaining utility on the retained classes. Existing class-level unlearning methods have achieved promising results under conventional evaluation metrics, including retain-set accuracy, forget-set accuracy, and unlearning time. These metrics measure whether the unlearned model still performs well on retained data, fails to recognize forgotten data, and completes the unlearning process efficiently~\cite{chundawat2023can, zhou2025decoupled, chundawat2023zero}.

However, these external metrics provide limited insight into how the forgetting effect is achieved internally. A model may obtain near-zero accuracy on the forgotten classes not because it has removed the internal representations associated with those classes, but because it has suppressed their output logits at the final classification layer. This distinction is important for class-level unlearning. If the forgetting effect is mainly realized by manipulating the final classification head, the model may only appear to forget under conventional metrics while still preserving class-related representations in earlier layers. Moreover, such manipulation may leave abnormal parameter patterns that reveal which classes have been forgotten. Therefore, evaluating class-level unlearning only through retain accuracy, forget accuracy, and time may overestimate the reliability and privacy of the unlearning process.

In this paper, we investigate this issue from the perspective of classification-head bias. For a standard neural classifier, the final linear classification head maps learned features to class logits through class-specific weights and biases. The bias term of each class can be regarded as a class-dependent baseline preference. If the bias of a particular class is shifted toward a sufficiently negative value, the corresponding logit can be suppressed across inputs, making the model unlikely to predict that class even when the learned feature representation remains unchanged. Motivated by this observation, we systematically analyze the role of classification-head bias in class-level unlearning. We show that, under softmax cross-entropy training, the gradient update direction tends to increase the bias of the ground-truth class and decrease the biases of non-ground-truth classes. Consequently, when a class is absent from the optimization data, as in retain-set-only fine-tuning or retraining, its corresponding bias tends to be continuously suppressed. This provides a theoretical explanation for why class-level unlearning methods may heavily rely on reducing forgotten-class biases.

To further expose this phenomenon, we introduce BiasShift as a diagnostic baseline. BiasShift directly subtracts a constant value from the biases of the forgotten-class classification heads. Although this operation is extremely simple and does not involve iterative optimization, it can achieve high retain-set accuracy, near-zero forget-set accuracy, and negligible unlearning time under conventional evaluation metrics. This result does not suggest that BiasShift is a desirable privacy-preserving unlearning method. Instead, it reveals a bias-dominated shortcut in conventional class-level unlearning evaluation: a model can satisfy standard metrics by suppressing the final-layer output probability of the forgotten classes, without necessarily removing their internal influence. More importantly, BiasShift leaves an obvious abnormal bias pattern, where the forgotten-class biases become much smaller than those of the retained classes. An attacker with access to the classification head can therefore infer the forgotten labels by inspecting the bias vector.

Based on the above analysis, this paper argues that reliable class-level unlearning should not only reduce the prediction accuracy on forgotten classes, but also avoid excessive dependence on abnormal forgotten-class bias suppression. To this end, we propose two bias-aware parameter redistribution mechanisms: Two-Stage Bias Gradient Reversal Mechanism (TS-BGRM) and Lower-Bound Hinge Regularization (LB-HR). TS-BGRM adopts a destroy-then-repair strategy. It first disrupts the decision behavior associated with the forgotten classes through bias-gradient reversal and then restores the utility on retained classes. This design aims to reduce the separation between forgotten-class and retained-class biases while maintaining unlearning effectiveness. LB-HR introduces a lower-bound hinge regularization term into the loss function to prevent forgotten-class biases from drifting toward extreme negative values during unlearning. Both mechanisms are designed to mitigate excessive bias dependence and redistribute the unlearning effect from extreme bias suppression to broader parameter adjustments.

Furthermore, we propose three bias-oriented evaluation metrics to complement conventional unlearning metrics. The Bias Stability Coefficient (BSC) measures the average bias deviation between forgotten and retained classes. The Median Bias Gap (MBG) evaluates whether the median forgotten-class bias remains comparable to the retained-class bias distribution. The Minimal Bias Score (MBS) further captures whether the most suppressed forgotten-class bias becomes an outlier. These metrics are intended to quantify the degree of bias dependence and the potential forgotten-label leakage risk caused by abnormal classification-head bias patterns. Together with retain-set accuracy, forget-set accuracy, and unlearning time, they provide a more comprehensive evaluation of class-level unlearning methods.

The main contributions of this paper are summarized as follows.

\begin{itemize}
    \item We provide a theoretical and empirical analysis of classification-head bias dynamics in class-level machine unlearning. Specifically, we show that under softmax cross-entropy training, the gradient update direction tends to increase the bias of the ground-truth class and decrease the biases of non-ground-truth classes. This explains why retain-set-only optimization can suppress the classification-head biases of absent, i.e., forgotten, classes.

    \item We introduce BiasShift as a diagnostic baseline to expose a bias-dominated shortcut in conventional class-level unlearning evaluation. By directly shifting the forgotten-class biases, BiasShift can achieve high retain-set accuracy, near-zero forget-set accuracy, and negligible unlearning time. This result demonstrates that conventional metrics alone are insufficient to determine whether the influence of forgotten classes has been reliably removed.

    \item We propose two bias-aware parameter redistribution mechanisms, namely Two-Stage Bias Gradient Reversal Mechanism (TS-BGRM) and Lower-Bound Hinge Regularization (LB-HR), to mitigate excessive forgotten-class bias suppression. TS-BGRM reduces bias separation through a destroy-then-repair optimization strategy, while LB-HR explicitly prevents forgotten-class biases from drifting toward extreme negative values.

    \item We design three bias-oriented evaluation metrics, including Bias Stability Coefficient (BSC), Median Bias Gap (MBG), and Minimal Bias Score (MBS), to quantify the dependence of unlearning methods on classification-head biases. These metrics complement conventional retain accuracy, forget accuracy, and unlearning time by revealing potential forgotten-label leakage from internal bias patterns.

    \item We conduct extensive experiments on CIFAR-10, CIFAR-100, and Tiny-ImageNet with multiple representative unlearning baselines, including Fine-tuning, NegGrad+, Random-label, SalUn, SCRUB, UNSIR, and SSD. The results verify the existence of the bias-level shortcut and show that TS-BGRM and LB-HR achieve competitive unlearning performance while producing more stable and less revealing bias distributions.
\end{itemize}

The remainder of this paper is organized as follows. Section~\ref{sec:related} reviews related studies on machine unlearning. Section~\ref{sec:motivation} presents the motivation and formulates the class-level unlearning problem. Section~\ref{sec:overview} provides an overview of the proposed framework. Section~\ref{sec:methodology} analyzes the bias dynamics of classification heads and introduces BiasShift, TS-BGRM, and LB-HR. Section~\ref{sec:experiments} reports the experimental results and analysis. Finally, Section~\ref{sec:conclusion} concludes this paper.

\section{Related Work}
\label{sec:related}

\subsection{Machine Unlearning}

Machine unlearning aims to remove the influence of specified training data from a trained model without retraining the model from scratch. It has attracted increasing attention due to privacy regulations, data correction requirements, and the need to eliminate low-quality or harmful training data. Existing machine unlearning methods can be broadly divided into exact unlearning and approximate unlearning.

Exact unlearning seeks to ensure that the unlearned model is equivalent, or sufficiently close, to a model retrained from scratch on the remaining data. A representative framework is SISA, which partitions the training set into multiple disjoint shards and trains independent sub-models on these shards. When a deletion request arrives, only the affected shard needs to be retrained, thereby reducing the cost compared with full retraining~\cite{bourtoule2021machine, yan2022arcane}. Such data-partitioning-based strategies provide a principled way to reduce retraining overhead, but their effectiveness often depends on pre-designed training structures and may be less flexible when applied to already trained models.

Approximate unlearning relaxes the strict equivalence requirement and instead aims to efficiently reduce the influence of the target data while preserving model utility. Existing approximate methods have explored diverse mechanisms, including influence-function-based approximation~\cite{koh2017understanding, dwork2006our, wu2022puma}, Fisher-information-based forgetting~\cite{golatkar2020eternal,golatkar2021mixed}, Hessian-based estimation~\cite{zheng2026accurate}, and sparsity-aware parameter modification~\cite{jia2023model}. These methods reduce the computational cost of unlearning by avoiding full retraining, but they usually evaluate unlearning mainly through external prediction behavior, such as accuracy on the retained and forgotten data. As a result, they may provide limited evidence regarding the internal mechanism through which forgetting is achieved.

\subsection{Class-Level Machine Unlearning}

Class-level machine unlearning focuses on removing all training samples belonging to one or more specified classes. Compared with sample-level unlearning, class-level unlearning imposes a stronger forgetting objective because the model is expected to lose its predictive capability for an entire semantic category while maintaining performance on the remaining classes. This setting is widely used in recent studies because it provides a clear and measurable way to evaluate whether the unlearned model can suppress the target classes.

Several representative methods have been proposed for class-level or class-centric unlearning. Gradient-ascent-based methods, such as NegGrad and its variants, directly optimize the forget set in the opposite direction to reduce the model's confidence on forgotten samples, often combined with retention objectives to prevent excessive degradation on retained classes~\cite{choi2023towards}. Random-label-based methods assign incorrect or random labels to forget samples, forcing the model to fit noisy supervision and thereby weakening its predictive ability on the forgotten classes~\cite{golatkar2020eternal}. Saliency-based unlearning methods, such as SalUn, identify parameters that are more relevant to the forget set and selectively update them to improve unlearning efficiency and stability~\cite{fan2023salun}. Distillation-based methods, such as SCRUB, formulate unlearning as a selective knowledge distillation problem, encouraging the unlearned model to deviate from the teacher model on the forget set while preserving similar behavior on the retain set~\cite{kurmanji2023towards}. Noise-based methods, such as UNSIR, first generate error-maximizing noise to impair the model components associated with forgotten classes and then repair the model using retained data~\cite{tarun2023fast}. Parameter-dampening methods, such as SSD, estimate parameter importance using Fisher information and selectively dampen parameters that are more related to the forget set, without requiring iterative gradient updates~\cite{foster2024fast}.

Beyond image classification, machine unlearning has also been extended to random forests~\cite{brophy2021machine}, federated learning~\cite{wang2022federated, liu2025blockful}, graph neural networks~\cite{xiao2017fashion, chen2022graph}, pre-trained models~\cite{li2024single,yao2024large}, and large models~\cite{yao2024machine, liu2025rethinking}. These studies demonstrate the broad applicability of unlearning techniques across different model architectures and learning paradigms. However, most existing class-level unlearning methods are still primarily evaluated using retain-set accuracy, forget-set accuracy, and unlearning time. Although these metrics are useful for measuring external prediction behavior and computational efficiency, they do not directly reveal whether the forgetting effect is achieved by removing class-related internal knowledge or by exploiting shortcuts in the final prediction layer.

\subsection{Internal Mechanisms and Bias-Oriented Evaluation}

Recent studies have started to question whether low forget-set accuracy necessarily indicates reliable unlearning. Some works suggest that an unlearned model may still preserve internal representations or residual information associated with the forgotten data, even when its external predictions appear to satisfy conventional unlearning criteria~\cite{gao2026illusion}. This observation indicates that evaluating unlearning solely from output behavior may be insufficient. For class-level unlearning, the final classification head is particularly important because it directly maps learned representations to class logits. If the output probability of a forgotten class is suppressed mainly through the final classification head, the model may appear to forget the class without fully eliminating its internal representations.

Closely related to this issue, recent work has observed that class-level unlearning may heavily affect the bias terms of the final classification layer~\cite{hatami2026class}. In particular, the biases corresponding to forgotten classes can become significantly smaller than those of retained classes, suggesting that the model may suppress forgotten classes through a bias-level shortcut. Such a shortcut is problematic for two reasons. First, it challenges the reliability of conventional unlearning metrics, since near-zero forget-set accuracy may be achieved by manipulating the final-layer bias rather than by removing the learned influence of the forgotten classes. Second, it may introduce a forgotten-label leakage risk, because the forgotten classes can become inferable from abnormal bias values in the classification head.

Different from existing methods that primarily aim to improve retain accuracy, forget accuracy, or unlearning efficiency, this paper focuses on the role of classification-head bias in class-level unlearning. We theoretically analyze the gradient dynamics of classification-head biases and show why retain-set-only optimization tends to suppress the biases of absent classes. Based on this analysis, we introduce BiasShift as a diagnostic baseline to expose the bias-dominated shortcut behind conventional evaluation. We further propose TS-BGRM and LB-HR to mitigate excessive forgotten-class bias suppression, and design BSC, MBG, and MBS as bias-oriented metrics to complement conventional unlearning evaluation. In this way, our work provides a parameter-level perspective for understanding, diagnosing, and reducing bias dependence in class-level machine unlearning.

\section{Motivation and Problem Formulation}
\label{sec:motivation}

This section presents the motivation and formal problem setting of this work. 
We first show that class-level forgetting can be largely controlled by the bias terms of the final classification head. 
This observation reveals a bias-dominated shortcut: a model may achieve near-zero accuracy on the forgotten classes by suppressing their output logits, without necessarily removing the internal representations associated with these classes. 
We then formulate the class-level unlearning problem and clarify why reliable unlearning should consider not only external prediction behavior but also abnormal parameter-level evidence.

\subsection{Motivation: Bias-Dominated Class Forgetting}
\label{subsec:motivation_bias}

For a standard neural classifier, the final linear classification head maps the learned feature representation to class logits. 
Besides the class-specific weight vectors, the bias terms act as class-dependent baseline preferences. 
For an input sample $x$, the logit of class $k$ is usually computed as
\begin{equation}
    z_k(x) = w_k^{\top}\phi(x) + b_k,
\end{equation}
where $\phi(x)$ denotes the feature representation, and $w_k$ and $b_k$ are the weight vector and bias term of the $k$-th classification head, respectively. 
Although the weight vector interacts with the input feature, the bias term directly shifts the logit of a class in an input-independent manner. 
Therefore, if the bias of a class is shifted toward a sufficiently negative value, the corresponding logit can be consistently suppressed across different inputs, making the model unlikely to predict this class even when its feature representation remains unchanged.

This property raises an important question for class-level machine unlearning: can a model appear to forget an entire class simply by modifying the bias term of the corresponding classification head? 
To examine this question, we consider a simple bias manipulation operation, referred to as \emph{BiasShift}. 
Given a trained model and a forgotten-class set $\mathcal{V}$, BiasShift directly subtracts a positive constant $\beta$ from the bias terms of the forgotten-class heads:
\begin{equation}
    b_c \leftarrow b_c - \beta, \quad c \in \mathcal{V},
\end{equation}
where $\beta>0$ controls the strength of logit suppression. 
This operation does not update the feature extractor or the classification weights. 
It only changes a small number of final-layer bias parameters.

\begin{figure}[!t]
    \centering
    \includegraphics[width=0.8\linewidth]{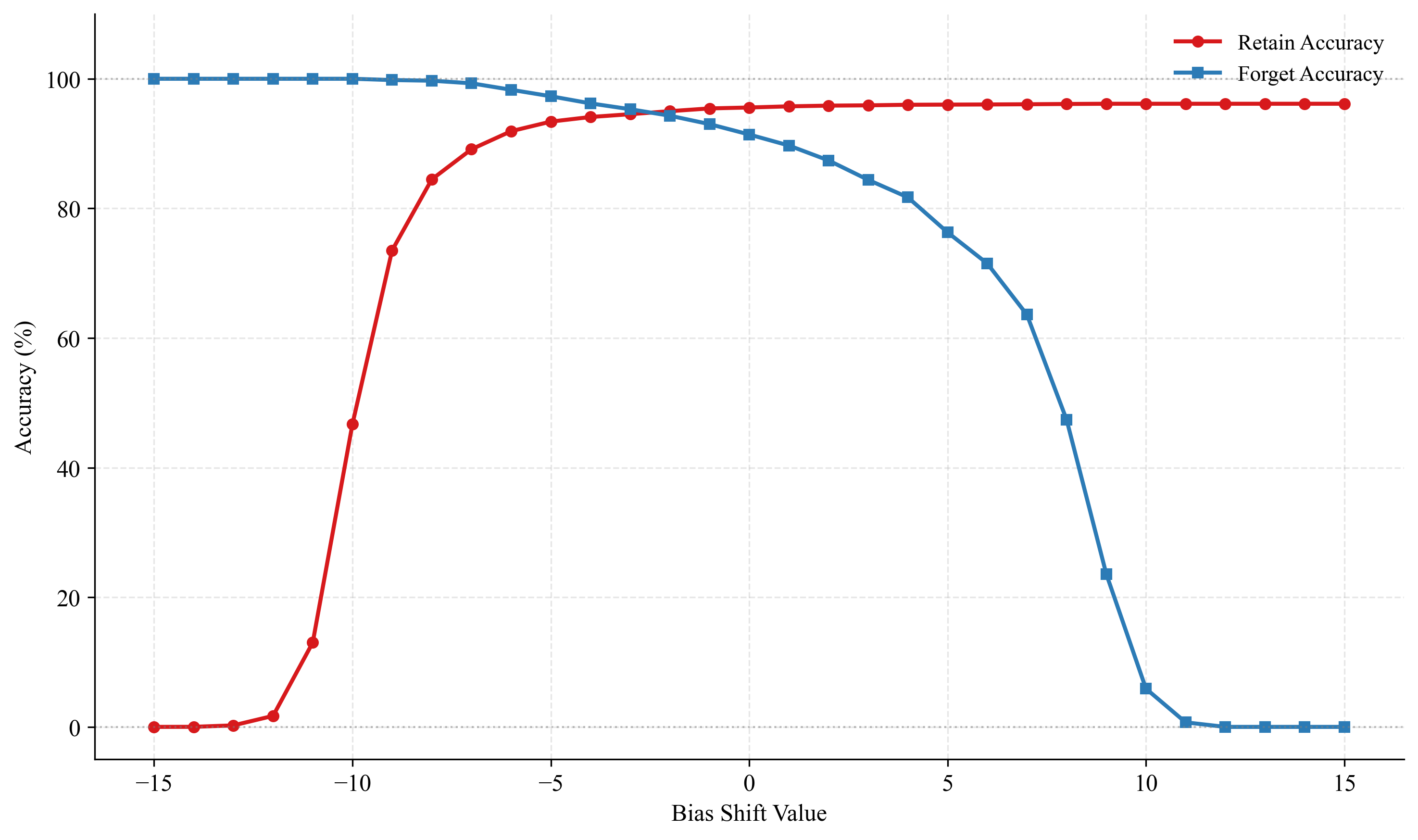} 
\caption{Effect of manually shifting the bias of the 5th-class classification head on CIFAR-10 using ResNet18.  A sufficiently negative bias shift suppresses predictions of the forgotten class and reduces the forget-set accuracy to zero, while an excessively positive shift causes the model to over-predict the forgotten class.}
    \label{fig:1}
\end{figure}

As shown in Fig.~\ref{fig:1}, when the bias of the forgotten-class head is progressively decreased, the accuracy on the forget set drops rapidly and eventually reaches zero. 
Meanwhile, the retain-set accuracy remains stable and may even slightly increase, because retained samples become less likely to be misclassified into the forgotten class. 
Conversely, increasing the forgotten-class bias produces the opposite effect: the model develops an excessive prior preference for the forgotten class and may incorrectly classify many inputs into this class. 
These observations indicate that the final-layer bias alone can substantially control the apparent class-level forgetting behavior of the model.

\begin{table}[!t]
    \centering
    \setlength{\tabcolsep}{3pt}
    \caption{Effect of the BiasShift method on ResNet18 for forgetting tasks on CIFAR10, CIFAR100, and Tiny-ImageNet datasets, with the number of forgotten categories being 1 and 3.}
    \label{tab:1}
    \begin{tabular}{l l c c c c c c c c}
        \toprule
        \multirow{2}{*}{\makecell{Dataset}} 
        & \multirow{2}{*}{\makecell{Method}} 
        & \multicolumn{4}{c}{1 class} 
        & \multicolumn{4}{c}{3 classes} \\
        \cmidrule(lr){3-6} \cmidrule(lr){7-10}
        & & $\beta$ & Retain & Forget & Time & $\beta$ & Retain & Forget & Time \\
        \midrule
        \multirow{2}{*}{CIFAR10} 
        & Origin & -- & 95.53 & 93.20 & -- & -- & 96.20 & 92.70 & -- \\
        & Unlearn & 15 & 96.14 & 0.00 & 0.018 & 15 & 97.21 & 0.00 & 0.016 \\
        \multirow{2}{*}{CIFAR100} 
        & Origin & -- & 76.49 & 76.00 & -- & -- & 76.00 & 77.41 & -- \\
        & Unlearn & 25 & 76.56 & 0.00 & 0.020 & 20 & 77.73 & 0.00 & 0.016 \\
        \multirow{2}{*}{\makecell{Tiny-\\ImageNet}} 
        & Origin & -- & 65.18 & 52.0 & -- & -- & 65.06 & 52.66 & -- \\
        & Unlearn & 10 & 65.26 & 0.00 & 0.027 & 15 & 65.42 & 0.00 & 0.030 \\
        \bottomrule
    \end{tabular}
\end{table}

Table~\ref{tab:1} further shows that this phenomenon consistently appears across different datasets and forgetting settings. 
By manually reducing the forgotten-class biases, BiasShift can drive the forget-set accuracy to zero while maintaining competitive retain-set accuracy. 
Since the operation only modifies a few bias entries, its computational cost is almost negligible compared with iterative unlearning methods or retraining. 
However, this result should not be interpreted as evidence that BiasShift is a reliable or privacy-preserving unlearning method. 
Instead, it exposes a limitation of conventional class-level unlearning evaluation: retain-set accuracy, forget-set accuracy, and unlearning time can be satisfied through a simple output-layer shortcut.

\begin{figure}[!t]
    \centering
    \includegraphics[width=0.8\linewidth]{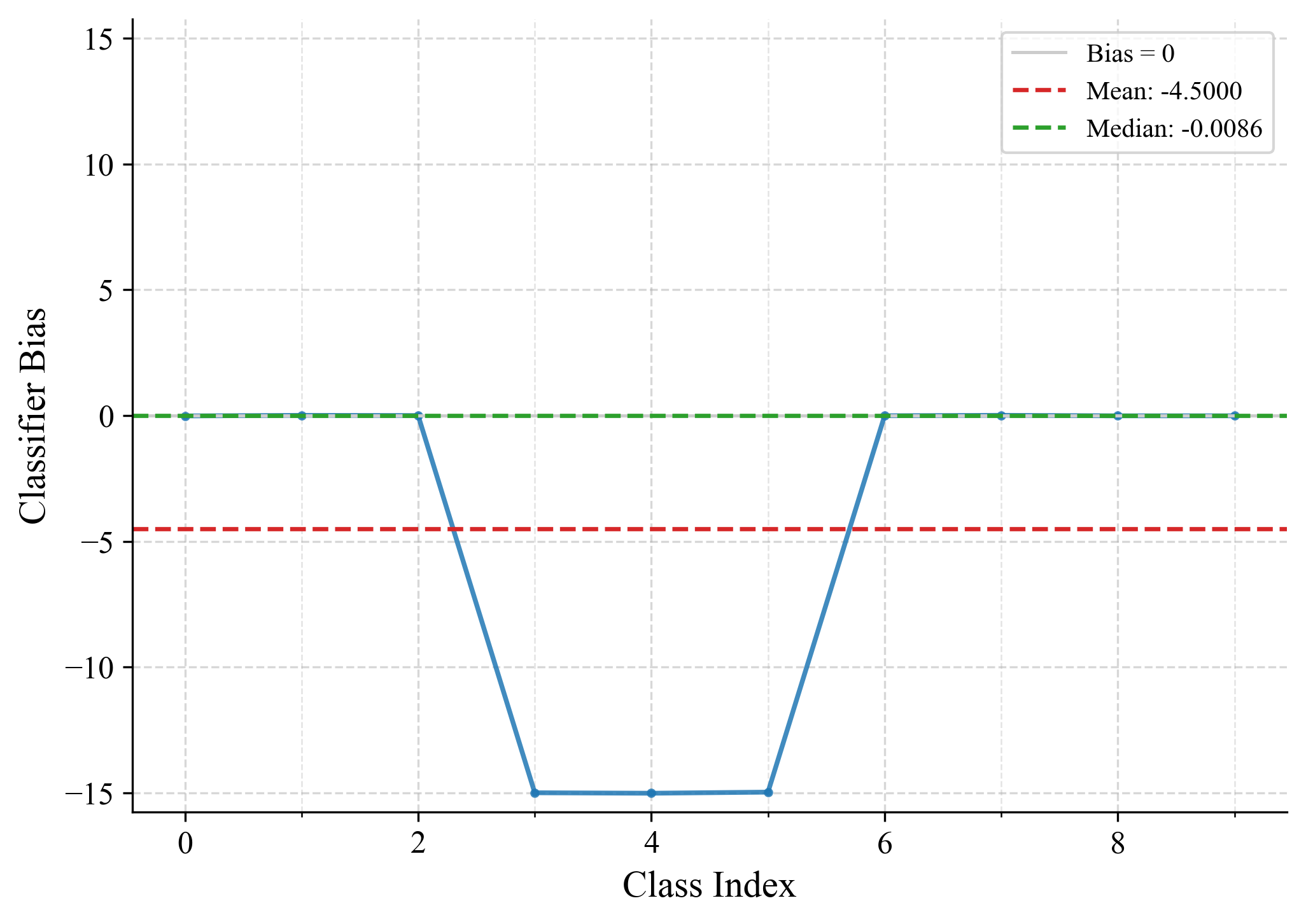} 
    \caption{Magnitude of the bias terms after applying the BiasShift method on ResNet18 over CIFAR10, where classes 3, 4, and 5 are the forgotten class indices.}
    \label{fig:2}
\end{figure}

The shortcut introduced by BiasShift also leads to a clear parameter-level leakage risk. 
As illustrated in Fig.~\ref{fig:2}, after the forgotten-class biases are manually reduced, they become significantly smaller than the biases of retained classes. 
An attacker with access to the final classification head can therefore infer the forgotten labels by inspecting the abnormal bias values. 
In this sense, classification-head bias is not merely an implementation detail, but a potential leakage channel in class-level unlearning. 
A method that achieves near-zero forget-set accuracy by pushing forgotten-class biases to extreme negative values may still leave strong evidence about which classes have been forgotten.

The above observations motivate the central question studied in this paper: how can we distinguish genuine class-level unlearning from bias-dominated output suppression? 
To answer this question, we need to go beyond conventional external metrics and analyze the internal parameter behavior of unlearned models. 
In particular, a reliable class-level unlearning method should not only reduce the model's predictive capability on forgotten classes and preserve utility on retained classes, but also avoid producing abnormal bias patterns that make the forgotten labels easily inferable. 
We next formalize the class-level unlearning setting and then develop a bias-aware analysis and mitigation framework based on this motivation.

\begin{figure}[!t]
    \centering
    \includegraphics[width=0.8\linewidth, keepaspectratio]{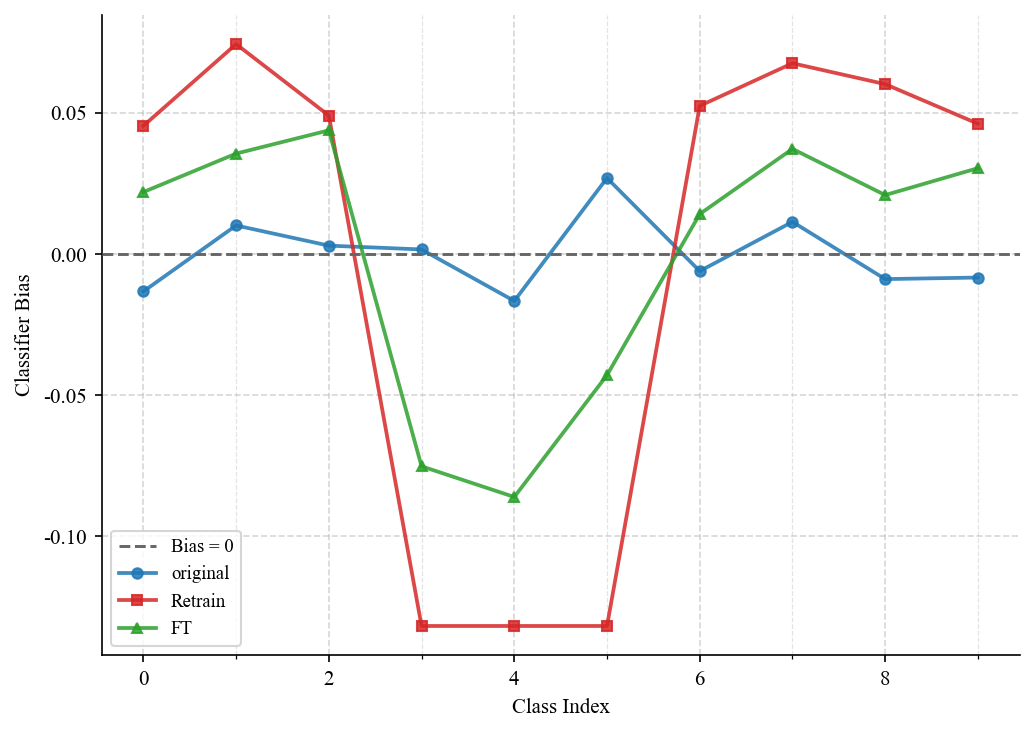} 
    \caption{Bias results of the model classification head trained on the CIFAR10 dataset using ResNet18. "Original" refers to the model trained on the full dataset $D$, "Retrain" denotes a new model trained solely on the retain set, and "FT" (Fine-Tuning) indicates continued training on the retain set based on the Original model. The forgotten classes are set to 3, 4, and 5.}
    \label{fig:bias_analysis}
\end{figure}

In addition to the manual BiasShift operation, we also observe similar bias suppression in standard class-level unlearning procedures. 
Fig.~\ref{fig:bias_analysis} compares the classification-head biases of the original model, the retrained model, and the fine-tuned model on CIFAR-10 with ResNet18. 
The original model is trained on the full dataset, while the retrained and fine-tuned models are optimized only on the retain set after removing classes 3, 4, and 5. 
Compared with the original model, both retraining and fine-tuning make the biases of the forgotten-class heads significantly smaller than those of retained classes. 
This result indicates that forgotten-class bias suppression is not only caused by the artificial BiasShift operation, but can also naturally arise from retain-set-only optimization. 
It further motivates our theoretical analysis of classification-head bias dynamics in Section~\ref{subsec:bias_dynamics}.

\subsection{Problem Formulation}
\label{subsec:problem_formulation}

Let $\mathcal{D}=\{(x_i,y_i)\}_{i=1}^{N}$ denote the original training dataset, where $x_i$ is an input sample and $y_i \in \mathcal{C}$ is its class label. 
Here, $\mathcal{C}=\{1,2,\ldots,C\}$ denotes the complete label space with $C$ classes. 
In class-level machine unlearning, a subset of classes is specified to be forgotten. 
We denote the forgotten-class set as $\mathcal{V}\subset \mathcal{C}$ and the retained-class set as $\mathcal{R}=\mathcal{C}\setminus \mathcal{V}$. 
Accordingly, the original training dataset is divided into the forget set and the retain set:
\begin{equation}
    \mathcal{D}_{f}=\{(x_i,y_i)\in \mathcal{D}\mid y_i\in \mathcal{V}\},
\end{equation}
\begin{equation}
    \mathcal{D}_{r}=\{(x_i,y_i)\in \mathcal{D}\mid y_i\in \mathcal{R}\},
\end{equation}
where $\mathcal{D}=\mathcal{D}_{r}\cup \mathcal{D}_{f}$ and $\mathcal{D}_{r}\cap \mathcal{D}_{f}=\emptyset$.

Let $M_{\mathrm{origin}}$ be a model trained on the full dataset $\mathcal{D}$. 
The goal of class-level unlearning is to obtain an unlearned model $M_{\mathrm{unlearn}}$ by applying an unlearning algorithm $\mathcal{U}$ to $M_{\mathrm{origin}}$:
\begin{equation}
    M_{\mathrm{unlearn}}=\mathcal{U}(M_{\mathrm{origin}},\mathcal{D}_{r},\mathcal{D}_{f}).
\end{equation}
Ideally, $M_{\mathrm{unlearn}}$ should behave similarly to a reference model $M_{\mathrm{retrain}}$ trained from scratch only on the retain set $\mathcal{D}_{r}$, while avoiding the computational cost of full retraining. 
In conventional class-level unlearning evaluation, this objective is usually assessed by three types of metrics: retain-set accuracy, forget-set accuracy, and unlearning time. 
A desirable unlearning method is expected to maintain high accuracy on $\mathcal{D}_{r}$, achieve low accuracy on $\mathcal{D}_{f}$, and consume substantially less time than retraining.

However, as discussed in Section~\ref{subsec:motivation_bias}, these conventional metrics only characterize the external prediction behavior of the unlearned model. 
They do not indicate whether the forgetting effect is achieved by removing or redistributing the learned influence of forgotten classes, or merely by suppressing the logits of forgotten classes through the final classification head. 
Therefore, in this paper, we consider a stronger bias-aware reliability requirement for class-level unlearning.

Specifically, let the final linear classification head of a neural classifier be parameterized by the class-specific weight matrix $W=[w_1,\ldots,w_C]$ and the bias vector $b=[b_1,\ldots,b_C]$. 
For a given input $x$, the logit of class $k$ is computed as
\begin{equation}
    z_k(x)=w_k^{\top}\phi(x)+b_k,
\end{equation}
where $\phi(x)$ denotes the feature representation extracted by the backbone network. 
For an unlearned model, we denote the bias values of forgotten classes and retained classes as
\begin{equation}
    b_{\mathcal{V}}=\{b_c\mid c\in \mathcal{V}\}, 
    \quad
    b_{\mathcal{R}}=\{b_c\mid c\in \mathcal{R}\}.
\end{equation}
A bias-dominated unlearning method may achieve low forget-set accuracy by making $b_{\mathcal{V}}$ significantly smaller than $b_{\mathcal{R}}$, thereby suppressing the prediction probability of forgotten classes. 
Although such a model may satisfy conventional metrics, the abnormal separation between $b_{\mathcal{V}}$ and $b_{\mathcal{R}}$ can reveal the forgotten labels from the final classification head.

Based on this observation, the objective studied in this paper is not limited to conventional class-level unlearning performance. 
Instead, we aim to analyze and mitigate the following bias-dependent shortcut:
\begin{equation}
\begin{aligned}
&\mathrm{Acc}_{r}(M_{\mathrm{unlearn}})\uparrow,\quad
\mathrm{Acc}_{f}(M_{\mathrm{unlearn}})\downarrow,\quad
T_{\mathrm{unlearn}}\downarrow,\\
&\text{but}\quad b_{\mathcal{V}} \ll b_{\mathcal{R}} .
\end{aligned}
\end{equation}
where $\mathrm{Acc}_{r}$ and $\mathrm{Acc}_{f}$ denote the retain-set accuracy and forget-set accuracy, respectively, and $T_{\mathrm{unlearn}}$ denotes the unlearning time. 
This situation indicates that the unlearned model performs well under conventional metrics but may still expose a clear forgotten-class signature through its bias vector.
In other words, an unlearned model may simultaneously achieve high retain-set accuracy, low forget-set accuracy, and low unlearning time, while leaving an abnormal bias gap where the forgotten-class biases are much smaller than the retained-class biases.

Therefore, a reliable class-level unlearning method should satisfy two complementary requirements. 
First, it should preserve the conventional unlearning objectives, namely maintaining utility on retained classes, reducing predictive capability on forgotten classes, and avoiding the high cost of full retraining. 
Second, it should avoid excessive forgotten-class bias suppression, so that the forgotten-class biases do not become obvious outliers compared with retained-class biases. 
In the following sections, we develop a unified framework to analyze this bias-dependent shortcut, introduce BiasShift as a diagnostic baseline, and propose TS-BGRM and LB-HR to reduce the dependence of class-level unlearning on extreme bias suppression.

\begin{figure*}[t]
    \centering
    \includegraphics[width=0.85\textwidth]{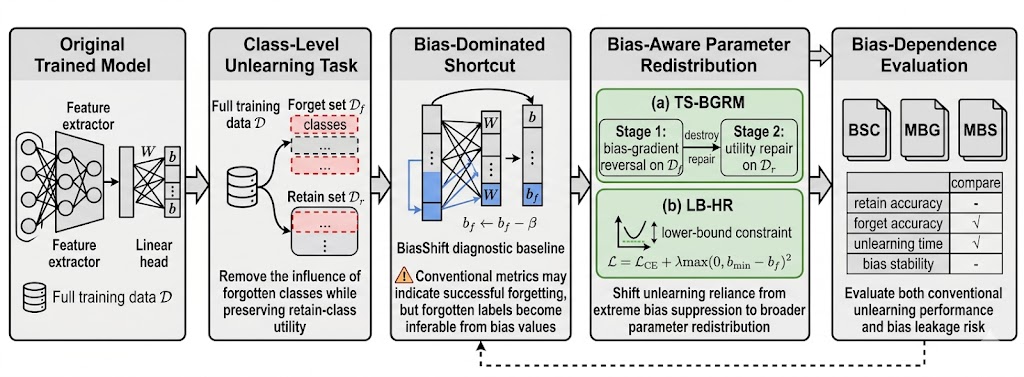}
    \caption{Overview of the proposed framework. BiasShift exposes a bias-dominated shortcut in class-level unlearning, while TS-BGRM and LB-HR mitigate excessive bias suppression through parameter redistribution. BSC, MBG, and MBS further quantify the bias dependence and leakage risk of unlearning methods.}
    \label{fig:overview}
\end{figure*}

\section{Overview of the Proposed Framework}
\label{sec:overview}

Based on the motivation and problem formulation in Section~\ref{sec:motivation}, this section provides an overview of the proposed framework. 
As illustrated in Fig.~\ref{fig:overview}, the framework consists of four components: bias-dominated shortcut identification, BiasShift-based diagnostic evaluation, bias-aware parameter redistribution, and bias-oriented assessment.

First, we identify a bias-dominated shortcut in class-level machine unlearning. 
Since the bias term of the final classification head shifts the class logit in an input-independent manner, decreasing the forgotten-class bias can directly suppress the prediction probability of that class. 
As a result, a model may achieve near-zero forget-set accuracy by output-layer suppression, without necessarily removing the internal representations associated with the forgotten class.

Second, we introduce BiasShift as a diagnostic baseline. 
BiasShift directly subtracts a positive constant from the forgotten-class biases and can satisfy conventional metrics, including retain-set accuracy, forget-set accuracy, and unlearning time. 
This result reveals that these metrics alone cannot distinguish reliable unlearning from bias-level shortcut behavior. 
Meanwhile, BiasShift leaves an abnormal bias pattern, making the forgotten labels inferable from the classification head. 
Therefore, it is used as a diagnostic tool rather than a privacy-preserving unlearning method.

Third, we propose two bias-aware parameter redistribution mechanisms, TS-BGRM and LB-HR, to mitigate excessive forgotten-class bias suppression. 
TS-BGRM adopts a destroy-then-repair strategy to disrupt forgotten-class decision behavior and restore retained-class utility, while LB-HR constrains forgotten-class biases from drifting toward extreme negative values. 
Both methods aim to maintain conventional unlearning performance while reducing abnormal bias separation between forgotten and retained classes.

Finally, we introduce three bias-oriented metrics, namely BSC, MBG, and MBS, to quantify the degree of bias dependence and potential forgotten-label leakage. 
Together with conventional metrics, they provide a more comprehensive evaluation of class-level unlearning from both external prediction behavior and internal parameter evidence.

The next section presents the theoretical analysis of classification-head bias dynamics and then details BiasShift, TS-BGRM, LB-HR, and the proposed bias-oriented metrics.

\section{Methodology}
\label{sec:methodology}

This section presents the methodology of this paper. 
We first analyze the gradient dynamics of the bias terms in the final linear classification head under softmax cross-entropy training. 
The analysis explains why optimization on the retain set naturally tends to suppress the biases of absent, i.e., forgotten, classes. 
Based on this result, we introduce BiasShift as a diagnostic baseline to expose the bias-dominated shortcut in conventional class-level unlearning evaluation. 
We then present two bias-aware mechanisms, TS-BGRM and LB-HR, which aim to reduce excessive forgotten-class bias suppression while maintaining unlearning effectiveness.

\subsection{Bias Dynamics in Classification Heads}
\label{subsec:bias_dynamics}

We begin by analyzing how the bias terms of the final classification head are updated during standard supervised training. 
Consider a neural classifier whose final linear head maps a feature representation $\phi(x)$ to class logits:
\begin{equation}
    z_k(x)=w_k^{\top}\phi(x)+b_k,\quad k\in\{1,\ldots,C\},
    \label{eq:logit_head}
\end{equation}
where $w_k$ and $b_k$ denote the weight vector and bias term of the $k$-th class, respectively. 
The softmax probability of class $k$ is given by
\begin{equation}
    p_k(x)=\frac{\exp(z_k(x))}
    {\sum_{j=1}^{C}\exp(z_j(x))}.
    \label{eq:softmax}
\end{equation}
For a training sample $(x,y)$, the cross-entropy loss is
\begin{equation}
    \mathcal{L}_{\mathrm{CE}}(x,y)=-\log p_y(x).
    \label{eq:ce_loss}
\end{equation}

For softmax cross-entropy, the gradient of the loss with respect to the logit $z_k$ is
\begin{equation}
    \frac{\partial \mathcal{L}_{\mathrm{CE}}}{\partial z_k}
    =
    p_k-\mathbb{I}(k=y),
    \label{eq:logit_gradient}
\end{equation}
where $\mathbb{I}(\cdot)$ is the indicator function. 
Since $\partial z_k / \partial b_k = 1$, the gradient with respect to the bias term $b_k$ is
\begin{equation}
    \frac{\partial \mathcal{L}_{\mathrm{CE}}}{\partial b_k}
    =
    p_k-\mathbb{I}(k=y).
    \label{eq:bias_gradient}
\end{equation}

Equation~\eqref{eq:bias_gradient} reveals a simple but important property of the classification-head bias update. 
For the ground-truth class $k=y$, we have
\begin{equation}
    \frac{\partial \mathcal{L}_{\mathrm{CE}}}{\partial b_y}
    =
    p_y-1 < 0,
    \label{eq:true_class_gradient}
\end{equation}
because $p_y\in(0,1)$. 
Under vanilla gradient descent with learning rate $\eta>0$, the update direction is therefore
\begin{equation}
    b_y \leftarrow b_y-\eta(p_y-1),
    \label{eq:true_class_update}
\end{equation}
which increases the bias of the ground-truth class for this single update. 
For any non-ground-truth class $k\neq y$, we have
\begin{equation}
    \frac{\partial \mathcal{L}_{\mathrm{CE}}}{\partial b_k}
    =
    p_k > 0,
    \label{eq:false_class_gradient}
\end{equation}
and its update becomes
\begin{equation}
    b_k \leftarrow b_k-\eta p_k,
    \label{eq:false_class_update}
\end{equation}
which decreases the corresponding bias for this single update.

This sign-determined update direction provides an explanation for the bias suppression phenomenon in class-level unlearning. 
Suppose that class $c$ is absent from the optimization data. 
Then $c$ never appears as the ground-truth label, and the bias gradient of class $c$ is always
\begin{equation}
    \frac{\partial \mathcal{L}_{\mathrm{CE}}}{\partial b_c}=p_c>0.
    \label{eq:absent_class_gradient}
\end{equation}
Under vanilla gradient descent, the corresponding update is
\begin{equation}
    b_c \leftarrow b_c-\eta p_c.
    \label{eq:absent_class_update}
\end{equation}
Therefore, when a class is completely absent from the optimization data, its bias is consistently updated toward smaller values. 
This is exactly the situation encountered by forgotten classes during retain-set-only fine-tuning or retraining. 
Since samples from forgotten classes are removed, the optimization process repeatedly treats these classes as non-ground-truth classes, leading to suppressed forgotten-class biases.

The gradient of the class-specific weight vector is
\begin{equation}
    \frac{\partial \mathcal{L}_{\mathrm{CE}}}{\partial w_k}
    =
    \bigl(p_k-\mathbb{I}(k=y)\bigr)\phi(x).
    \label{eq:weight_gradient}
\end{equation}
Unlike the bias gradient, the weight update is modulated by the feature representation $\phi(x)$ and does not have a globally fixed sign. 
Thus, the bias term provides a more direct and input-independent channel for suppressing the output probability of a class. 
This distinction helps explain why class-level unlearning methods that optimize on retained data may achieve apparent forgetting by excessively reducing the forgotten-class biases.

It should be noted that the above analysis characterizes the update direction under standard softmax cross-entropy and vanilla gradient descent. 
In practical training, mini-batch composition, optimizers, weight decay, and additional regularization terms may affect the exact update trajectory. 
Nevertheless, the sign structure in Eq.~\eqref{eq:bias_gradient} explains a fundamental tendency: if a class is absent from the optimization data, its classification-head bias receives no positive ground-truth update and is repeatedly pushed downward by non-ground-truth updates. 
This motivates us to use classification-head bias as a key perspective for diagnosing class-level unlearning.

\subsection{BiasShift as a Diagnostic Baseline}
\label{subsec:biasshift}

The bias dynamics above suggest that suppressing the bias of a forgotten class can directly reduce its prediction probability. 
To examine how much of class-level forgetting can be achieved through this output-layer shortcut, we introduce BiasShift as a diagnostic baseline. 
BiasShift is intentionally simple: it directly modifies the bias terms of the forgotten-class heads without updating the feature extractor or the classification weights.

Let $\mathcal{V}$ denote the set of forgotten classes. 
Given an original model $M_{\mathrm{origin}}$ with classification-head bias vector $b=[b_1,\ldots,b_C]$, BiasShift performs the following operation:
\begin{equation}
    b_c \leftarrow b_c-\beta,\quad c\in\mathcal{V},
    \label{eq:biasshift}
\end{equation}
where $\beta>0$ is a constant controlling the suppression strength. 
For classes not in $\mathcal{V}$, the corresponding bias terms remain unchanged. 
Since Eq.~\eqref{eq:biasshift} only modifies a small number of scalar parameters, BiasShift has negligible computational cost and does not require iterative optimization.

The effect of BiasShift can be understood directly from the logit expression in Eq.~\eqref{eq:logit_head}. 
For a forgotten class $c\in\mathcal{V}$, the shifted logit becomes
\begin{equation}
    z_c'(x)=w_c^{\top}\phi(x)+b_c-\beta
    =
    z_c(x)-\beta.
    \label{eq:shifted_logit}
\end{equation}
Thus, increasing $\beta$ uniformly lowers the logit of the forgotten class for all inputs. 
When $\beta$ is sufficiently large, the forgotten-class probability after softmax becomes very small, making the model unlikely to predict any input as the forgotten class. 
This explains why BiasShift can drive the forget-set accuracy to zero while preserving the learned feature extractor and most of the classification head.

However, the purpose of BiasShift is not to provide a privacy-preserving unlearning solution. 
Instead, it serves as a diagnostic baseline for evaluating the reliability of conventional class-level unlearning metrics. 
If a model can obtain high retain-set accuracy, near-zero forget-set accuracy, and negligible unlearning time merely by shifting forgotten-class biases, then these metrics alone cannot distinguish reliable unlearning from output-layer suppression. 
BiasShift therefore exposes a bias-dominated shortcut in existing evaluation protocols.

Moreover, BiasShift leaves an obvious parameter-level signature. 
Because the forgotten-class biases are directly shifted toward much smaller values, they can become clear outliers compared with the retained-class biases. 
An attacker with access to the final classification head may infer the forgotten labels by inspecting the bias vector. 
Therefore, BiasShift demonstrates both the effectiveness and the risk of bias-dominated forgetting: it can satisfy conventional metrics, but it also reveals that such metrics may overlook forgotten-label leakage through abnormal classification-head bias patterns.

For this reason, BiasShift is used throughout this paper as a diagnostic tool rather than as the final unlearning method. 
The following subsections introduce two bias-aware mechanisms, TS-BGRM and LB-HR, which are designed to maintain competitive unlearning performance while reducing excessive forgotten-class bias suppression.

\subsection{Bias-Aware Parameter Redistribution}
\label{subsec:bias_aware_redistribution}

The analysis above shows that classification-head bias provides an input-independent shortcut for suppressing the prediction probability of forgotten classes. 
Although such suppression can effectively reduce forget-set accuracy, it may also produce abnormal bias patterns that reveal the forgotten labels. 
Therefore, a reliable class-level unlearning method should avoid concentrating the forgetting effect only on the extreme reduction of forgotten-class biases.

To this end, we propose a bias-aware parameter redistribution perspective. 
The goal is not to completely eliminate the role of bias terms in unlearning, since the final classification head inevitably participates in class prediction. 
Instead, we aim to reduce excessive dependence on forgotten-class bias suppression and encourage the unlearning effect to be redistributed to broader classification-head parameters, especially the class-specific weight vectors. 
In this way, the model can still lose its predictive capability on forgotten classes, but the forgotten-class biases are less likely to become extreme outliers compared with retained-class biases.

In this paper, we focus on the final classification head and freeze the feature extractor $\phi(\cdot)$ during the proposed bias-aware unlearning procedures. 
This design has two motivations. 
First, it allows us to directly study how the final-layer weights and biases contribute to class-level forgetting, without introducing additional variations from the backbone feature extractor. 
Second, it improves computational efficiency, because only a small number of classification-head parameters need to be updated. 
It should be emphasized that freezing $\phi(\cdot)$ does not imply that the feature extractor is unimportant. 
Rather, it provides a controlled setting for analyzing whether the forgetting effect can be shifted away from extreme bias manipulation toward other trainable parameters in the classification head.

Based on this idea, we develop two bias-aware mechanisms. 
The first one, Two-Stage Bias Gradient Reversal Mechanism (TS-BGRM), uses a destroy-then-repair strategy to disrupt the decision behavior associated with forgotten classes and then restore the performance on retained classes. 
The second one, Lower-Bound Hinge Regularization (LB-HR), explicitly constrains the forgotten-class biases from drifting toward excessively negative values during optimization. 
Both mechanisms are designed to preserve conventional unlearning performance while reducing abnormal bias separation between forgotten and retained classes.

\begin{figure}[!t]
    \centering
    \includegraphics[width=1\linewidth, keepaspectratio]{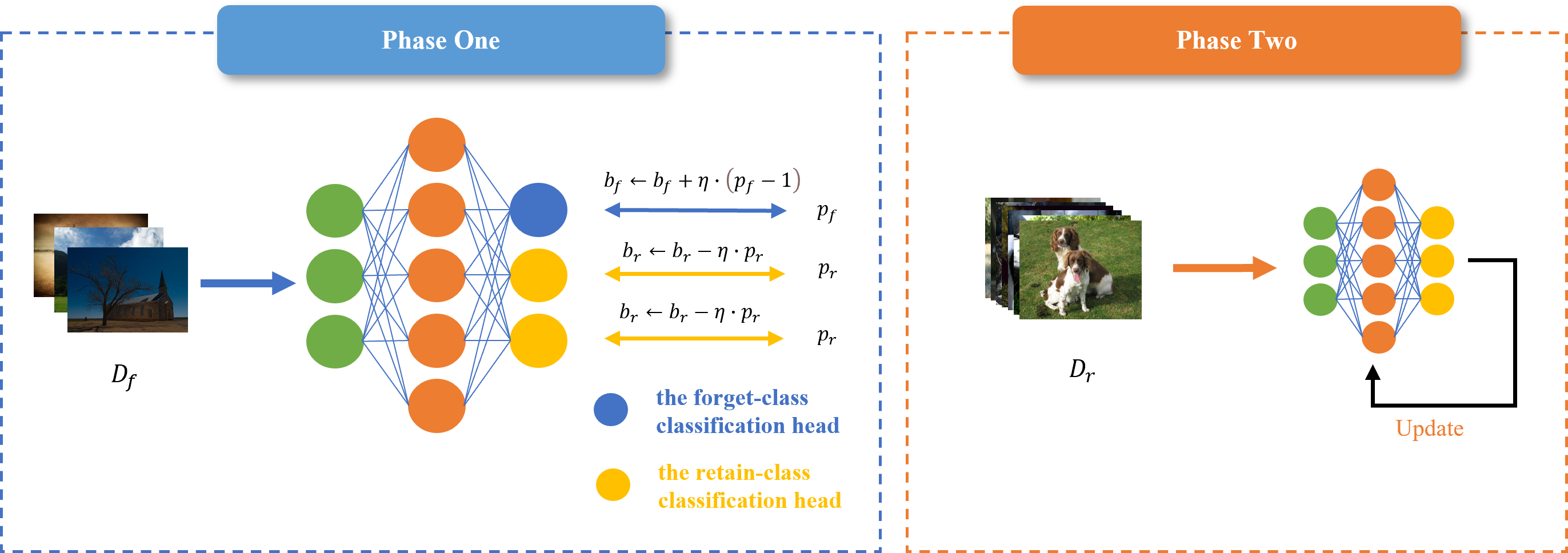} 
\caption{Illustration of the Two-Stage Bias Gradient Reversal Mechanism (TS-BGRM). 
The destroy stage uses the forget set with bias-gradient reversal to disrupt forgotten-class decision behavior, while the repair stage uses the retain set to recover retained-class utility.}
    \label{fig:tsbgrm}
\end{figure}

\subsection{Two-Stage Bias Gradient Reversal Mechanism}
\label{subsec:ts_bgrm}

We first introduce the Two-Stage Bias Gradient Reversal Mechanism (TS-BGRM). 
The key idea of TS-BGRM is to avoid the one-sided suppression of forgotten-class biases that commonly occurs in retain-set-only fine-tuning. 
Instead of directly optimizing only on the retain set, TS-BGRM separates the unlearning process into two stages: a destroy stage and a repair stage. 
The destroy stage weakens the model's decision behavior on forgotten classes, while the repair stage restores the utility on retained classes.

Let $\mathcal{V}$ and $\mathcal{R}$ denote the forgotten-class set and the retained-class set, respectively. 
In the first stage, TS-BGRM uses the forget set $\mathcal{D}_{f}$ to perform a short optimization process. 
Different from standard fine-tuning, we reverse the gradients of the bias terms corresponding to forgotten classes during backpropagation:
\begin{equation}
    \frac{\partial \widetilde{\mathcal{L}}}{\partial b_c}
    =
    \begin{cases}
    -\dfrac{\partial \mathcal{L}_{\mathrm{CE}}}{\partial b_c}, & c\in \mathcal{V},\\[6pt]
    \dfrac{\partial \mathcal{L}_{\mathrm{CE}}}{\partial b_c}, & c\in \mathcal{R}.
    \end{cases}
    \label{eq:bgrm_gradient}
\end{equation}
Here, $\mathcal{L}_{\mathrm{CE}}$ denotes the standard cross-entropy loss, and $\widetilde{\mathcal{L}}$ denotes the effective objective induced by the bias-gradient reversal operation. 
This operation is applied only to the bias terms of the classification head. 
Other trainable parameters in the classification head are updated according to the standard gradient.

The intuition behind Eq.~\eqref{eq:bgrm_gradient} is as follows. 
When the forget set is used for ordinary fine-tuning, the forgotten classes appear as ground-truth labels, and their biases tend to increase. 
This may reinforce the model's preference for forgotten classes and make subsequent unlearning more difficult. 
By reversing the forgotten-class bias gradients, TS-BGRM changes the bias update direction for forgotten classes and disrupts the model's original decision behavior on the forget set. 
At the same time, because retained classes do not appear as ground-truth labels in $\mathcal{D}_{f}$, their biases naturally receive non-ground-truth updates and tend to decrease. 
As a result, the destroy stage weakens the decision structure related to forgotten classes without simply making the forgotten-class biases much smaller than all retained-class biases.

More concretely, for a forget-set sample $(x,y)$ with $y\in\mathcal{V}$, the standard bias gradient of the ground-truth forgotten class is
\begin{equation}
    \frac{\partial \mathcal{L}_{\mathrm{CE}}}{\partial b_y}
    =
    p_y-1 < 0.
    \label{eq:forget_bias_standard_gradient}
\end{equation}
Under ordinary gradient descent, this negative gradient would increase $b_y$. 
With bias-gradient reversal, the effective gradient becomes
\begin{equation}
    \frac{\partial \widetilde{\mathcal{L}}}{\partial b_y}
    =
    1-p_y > 0,
    \label{eq:forget_bias_reversed_gradient}
\end{equation}
and the update direction becomes
\begin{equation}
    b_y \leftarrow b_y-\eta(1-p_y),
    \label{eq:forget_bias_reversed_update}
\end{equation}
which decreases the forgotten-class bias for this update. 
For a retained class $r\in\mathcal{R}$, since $r\neq y$, the standard gradient is
\begin{equation}
    \frac{\partial \mathcal{L}_{\mathrm{CE}}}{\partial b_r}=p_r>0,
    \label{eq:retain_bias_stage1_gradient}
\end{equation}
and its update is
\begin{equation}
    b_r \leftarrow b_r-\eta p_r.
    \label{eq:retain_bias_stage1_update}
\end{equation}
Thus, during the destroy stage, both forgotten-class and retained-class biases can be pushed downward, which helps avoid a large bias gap caused by suppressing only the forgotten-class heads.

After the destroy stage, the model's prediction behavior on forgotten classes is disrupted, but its performance on retained classes may also be affected. 
Therefore, the second stage performs repair using the retain set $\mathcal{D}_{r}$. 
In this stage, TS-BGRM applies standard fine-tuning without bias-gradient reversal. 
For a retained-class sample $(x,y)$ with $y\in\mathcal{R}$, the retained-class bias receives the standard ground-truth update:
\begin{equation}
    b_y \leftarrow b_y-\eta(p_y-1),
    \label{eq:repair_retain_update}
\end{equation}
which helps restore the model's utility on retained classes. 
For a forgotten class $c\in\mathcal{V}$, since it is absent from $\mathcal{D}_{r}$, its bias receives the non-ground-truth update:
\begin{equation}
    b_c \leftarrow b_c-\eta p_c.
    \label{eq:repair_forget_update}
\end{equation}
However, because the destroy stage has already weakened the model's confidence on forgotten classes, the probability $p_c$ is usually small during the repair stage. 
Therefore, the additional decrease of forgotten-class biases can be limited, reducing the risk that they become extreme negative outliers.

Overall, TS-BGRM differs from conventional retain-set-only fine-tuning in two aspects. 
First, it uses the forget set in the destroy stage to actively disrupt the decision behavior associated with forgotten classes, rather than relying only on the absence of forgotten-class samples in the retain set. 
Second, the bias-gradient reversal operation prevents the unlearning effect from being realized merely by one-sided forgotten-class bias suppression. 
This design encourages a more balanced bias distribution between forgotten and retained classes while maintaining low forget-set accuracy and high retain-set accuracy.

We also consider an ablation variant named Two-Stage Bias Gradient Mechanism (TS-BGM), which follows the same two-stage destroy-then-repair procedure but does not reverse the forgotten-class bias gradients in the first stage. 
The comparison between TS-BGM and TS-BGRM helps isolate the effect of bias-gradient reversal. 
Although the two variants may achieve similar retain-set and forget-set accuracies under conventional metrics, TS-BGRM is expected to produce more stable bias patterns, which can be captured by the proposed bias-oriented metrics in the experimental section.

\subsection{Lower-Bound Hinge Regularization for Bias Stability}
\label{LB-HR}

As a complementary mechanism to TS-BGRM, we introduce Lower-Bound Hinge Regularization (LB-HR) to explicitly prevent forgotten-class biases from drifting toward extreme negative values. 
In retain-set-only optimization, forgotten classes are absent from the training data, and their bias terms tend to be repeatedly updated as non-ground-truth classes. 
This may make the forgotten-class biases significantly smaller than retained-class biases, leaving an abnormal parameter-level signature. 
LB-HR addresses this issue by adding a lower-bound hinge penalty to the training objective:
\begin{equation}
    \mathcal{L}_{\mathrm{LB\text{-}HR}}
    =
    \mathcal{L}_{\mathrm{CE}}
    +
    \lambda
    \sum_{c\in\mathcal{V}}
    \left[\max(0,b_{\min}-b_c)\right]^2 ,
    \label{eq:lbhr_loss}
\end{equation}
where $\mathcal{V}$ is the forgotten-class set, $b_{\min}$ is the lower bound for forgotten-class biases, and $\lambda>0$ controls the regularization strength. 
The penalty is activated only when $b_c<b_{\min}$, and therefore it does not force the forgotten-class biases to increase unconditionally.

For a forgotten class $c\in\mathcal{V}$, the gradient of the regularization term is
\begin{equation}
    \frac{\partial \mathcal{L}_{\mathrm{reg}}}{\partial b_c}
    =
    \begin{cases}
    -2\lambda(b_{\min}-b_c), & b_c < b_{\min},\\[4pt]
    0, & b_c \geq b_{\min}.
    \end{cases}
    \label{eq:lbhr_reg_gradient}
\end{equation}
When $b_c<b_{\min}$, this gradient provides a positive correction under gradient descent:
\begin{equation}
    b_c^{(t+1)}
    =
    b_c^{(t)}
    -
    \eta
    \frac{\partial \mathcal{L}_{\mathrm{CE}}}{\partial b_c}
    +
    2\eta\lambda(b_{\min}-b_c^{(t)}).
    \label{eq:lbhr_update}
\end{equation}
Since $b_{\min}-b_c^{(t)}>0$, the last term pulls the forgotten-class bias upward once it falls below the lower bound. 
Thus, LB-HR counteracts the one-sided decrease of forgotten-class biases during retain-set optimization and prevents them from becoming extreme negative outliers.

Unlike BiasShift, which deliberately suppresses forgotten-class biases as a diagnostic operation, LB-HR aims to stabilize these biases while allowing the classification-head weights and other trainable parameters to contribute to the unlearning effect. 
Compared with TS-BGRM, which relies on a two-stage destroy-then-repair strategy, LB-HR provides a direct loss-level constraint for bias stability. 
Both mechanisms are designed to maintain conventional unlearning performance while reducing abnormal bias separation between forgotten and retained classes.

\section{Experiments}
\label{sec:experiments}

This section evaluates the proposed bias-aware class-level unlearning framework. 
The experiments are designed to answer the following research questions:

\begin{itemize}
    \item \textbf{RQ1:} Can a simple bias-level manipulation satisfy conventional class-level unlearning metrics?
    
    \item \textbf{RQ2:} Do existing class-level unlearning methods exhibit different degrees of dependence on forgotten-class bias suppression?
    
    \item \textbf{RQ3:} Can the proposed TS-BGRM and LB-HR methods reduce abnormal forgotten-class bias suppression while maintaining competitive retain-set utility and forgetting efficacy?
    
    \item \textbf{RQ4:} Do the proposed bias-oriented metrics provide complementary evidence beyond retain-set accuracy, forget-set accuracy, and unlearning time?
\end{itemize}


\subsection{Experimental Setting}
\label{subsec:experimental_setting}

\subsubsection{Evaluation Metrics}

We evaluate each unlearning method from two complementary perspectives: conventional unlearning performance and bias-oriented reliability.

\textbf{Conventional unlearning metrics.}
Following common practice in class-level machine unlearning, we first report retain-set accuracy, forget-set accuracy, and unlearning time. 
The retain-set accuracy measures the model utility on samples belonging to retained classes, while the forget-set accuracy measures the remaining predictive capability on samples belonging to forgotten classes. 
A desirable unlearning method should maintain high retain-set accuracy and achieve low forget-set accuracy. 
The unlearning time measures the computational cost required to obtain the unlearned model.

To make the time cost comparable across datasets and settings, we further report the Recovery Time Ratio (RTR), which normalizes the unlearning time by the time required for retraining from scratch on the retain set:
\begin{equation}
    \mathrm{RTR}
    =
    \frac{T_{\mathrm{unlearn}}}{T_{\mathrm{retrain}}}
    \times 100\%,
    \label{eq:rtr}
\end{equation}
where $T_{\mathrm{unlearn}}$ denotes the time consumed by an unlearning method, and $T_{\mathrm{retrain}}$ denotes the time required to retrain the model using only the retain set. 
A smaller RTR indicates higher unlearning efficiency. 
For BiasShift, we only measure the time required to modify the selected bias entries after the trained model is loaded, because BiasShift does not involve iterative optimization.

\textbf{Bias-oriented metrics.}
Conventional metrics only evaluate the external prediction behavior of the unlearned model. 
To measure whether the unlearning effect relies excessively on forgotten-class bias suppression, we introduce three bias-oriented metrics: Bias Stability Coefficient (BSC), Median Bias Gap (MBG), and Minimal Bias Score (MBS). 
Let $\mathcal{V}$ and $\mathcal{R}$ denote the forgotten-class set and retained-class set, respectively. 
Let $b_{\mathcal{V}}=\{b_c\mid c\in\mathcal{V}\}$ and $b_{\mathcal{R}}=\{b_c\mid c\in\mathcal{R}\}$ denote the corresponding bias values in the final classification head.

The Bias Stability Coefficient (BSC) measures the deviation between the mean forgotten-class bias and the mean retained-class bias:
\begin{equation}
    \mathrm{BSC}
    =
    \frac{1}
    {1+\left|\bar{b}_{\mathcal{V}}-\bar{b}_{\mathcal{R}}\right|}
    \times 100\%,
    \label{eq:bsc}
\end{equation}
where $\bar{b}_{\mathcal{V}}$ and $\bar{b}_{\mathcal{R}}$ are the average bias values of forgotten and retained classes, respectively. 
A larger BSC indicates a smaller average bias deviation between forgotten and retained classes.

However, the mean bias may hide extreme values when multiple classes are forgotten. 
Therefore, we further introduce the Median Bias Gap (MBG), which compares the median forgotten-class bias with the minimum retained-class bias:
\begin{equation}
    \mathrm{MBG}
    =
    \sigma
    \left(
    \mathrm{median}(b_{\mathcal{V}})
    -
    \min(b_{\mathcal{R}})
    \right)
    \times 100\%,
    \label{eq:mbg}
\end{equation}
where $\sigma(\cdot)$ denotes the sigmoid function. 
MBG evaluates whether the typical forgotten-class bias remains comparable to the lower range of retained-class biases.

The Minimal Bias Score (MBS) focuses on the most suppressed forgotten-class bias:
\begin{equation}
    \mathrm{MBS}
    =
    \sigma
    \left(
    \min(b_{\mathcal{V}})
    -
    \min(b_{\mathcal{R}})
    \right)
    \times 100\%.
    \label{eq:mbs}
\end{equation}
MBS is designed to capture whether any forgotten-class bias becomes an extreme negative outlier. 
When only one class is forgotten, MBG and MBS are equivalent.

For MBG and MBS, a value above $50\%$ indicates that the corresponding forgotten-class bias statistic is not smaller than the minimum retained-class bias. 
Therefore, higher BSC, MBG, and MBS values generally indicate more stable and less revealing bias patterns. 
These metrics do not replace conventional unlearning metrics; instead, they provide complementary evidence about whether an unlearned model leaves abnormal classification-head bias signatures.

\subsubsection{Datasets and Models}

We conduct experiments on three widely used image classification benchmarks: CIFAR-10, CIFAR-100, and Tiny-ImageNet. 
CIFAR-10 contains 60,000 color images from 10 classes, with 50,000 training images and 10,000 test images~\cite{krizhevsky2010cifar}. 
CIFAR-100 has the same number of images but contains 100 fine-grained classes. 
Tiny-ImageNet contains 200 classes, with RGB images resized to $64\times64$ pixels~\cite{xiao2017fashion}. 
These datasets provide different levels of classification granularity and difficulty, allowing us to evaluate whether the bias-dominated shortcut and the proposed mitigation methods are consistent across different class spaces.

Following common practice in image-based unlearning studies, we use ResNet18 as the backbone model for all datasets~\cite{he2016deep}. 
The model is first trained on the full training set to obtain $M_{\mathrm{origin}}$. 
For class-level unlearning, we consider both single-class and three-class forgetting settings. 
In the single-class setting, one target class is selected as the forgotten class. 
In the three-class setting, three classes are selected as forgotten classes. 
The remaining classes form the retained-class set. 
All methods are evaluated on the same retain and forget splits for fair comparison.

\subsubsection{Compared Methods}

We compare the proposed methods with representative class-level unlearning baselines.

\textbf{Retrain} trains a new model from scratch using only the retain set. 
It is usually regarded as the gold-standard reference for unlearning but requires the highest computational cost.

\textbf{Fine-tuning (FT)} continues training the original model on the retain set. 
It is a simple and efficient baseline, but it may suppress forgotten-class biases because forgotten classes are absent from the optimization data.

\textbf{NegGrad+} uses gradient ascent on the forget set together with a retention objective to reduce the model's confidence on forgotten samples while preserving retained-class performance.

\textbf{Random-label} assigns incorrect or random labels to forget samples, forcing the model to weaken its original decision behavior on forgotten classes.

\textbf{SalUn} updates parameters that are more relevant to the forget set, aiming to improve the efficiency and stability of the unlearning process.

\textbf{SCRUB} formulates unlearning as a selective knowledge distillation problem, encouraging deviation from the teacher model on forgotten data while maintaining consistency on retained data.

\textbf{UNSIR} uses error-maximizing noise to impair the model components related to forgotten classes and then repairs the model with retained data.

\textbf{SSD} estimates parameter importance and selectively dampens parameters associated with forgotten data without iterative gradient-based optimization.

In addition to these existing baselines, we include several methods related to our analysis.

\textbf{BiasShift} is used as a diagnostic baseline. 
It directly shifts the forgotten-class biases by a constant value and does not update other parameters. 
Its purpose is to expose the bias-dominated shortcut in conventional evaluation rather than to serve as a privacy-preserving final solution.

\textbf{Shallow Fine-tuning (SF)} freezes the feature extractor and fine-tunes only the final classification head. 
This baseline helps examine whether the effectiveness of the proposed methods is merely caused by updating fewer parameters.

\textbf{TS-BGM} is an ablation variant of TS-BGRM. 
It follows the same two-stage destroy-then-repair procedure but does not reverse the forgotten-class bias gradients in the destroy stage. 
The comparison between TS-BGM and TS-BGRM isolates the effect of bias-gradient reversal.

\textbf{TS-BGRM} and \textbf{LB-HR} are the two proposed bias-aware mitigation methods. 
TS-BGRM uses bias-gradient reversal in a two-stage optimization process, while LB-HR introduces a lower-bound hinge regularization term to stabilize forgotten-class biases.

All methods are evaluated using both conventional metrics and the proposed bias-oriented metrics. 
This allows us to compare not only their external unlearning performance but also their internal bias patterns.

\begin{table*}[!htbp]
    \centering
    \caption{Test accuracy of models on the retain set and forget set.}
    \label{tab:accuracy}
    \begin{tabular}{@{}lccccccccccccc@{}}
        \toprule
        \multirow{3}{*}{\makecell{Method}} 
        & \multicolumn{4}{c}{CIFAR10} 
        & \multicolumn{4}{c}{CIFAR100} 
        & \multicolumn{4}{c}{Tiny-ImageNet}\\
         \cmidrule(lr){2-5} \cmidrule(lr){6-9} \cmidrule(lr){10-13}
        & \multicolumn{2}{c}{1} & \multicolumn{2}{c}{3} & \multicolumn{2}{c}{1} & \multicolumn{2}{c}{3} & \multicolumn{2}{c}{1} & \multicolumn{2}{c}{3} \\
        \cmidrule(lr){2-3} \cmidrule(lr){4-5}  \cmidrule(lr){6-7} \cmidrule(lr){8-9} \cmidrule(lr){10-11} \cmidrule(lr){12-13}
        & Retain & Forget & Retain & Forget & Retain & Forget & Retain & Forget & Retain & Forget  & Retain & Forget \\
        \midrule
\rowcolor{gray!10}
        Original & 95.53 & 93.20 & 96.20 & 92.70 & 76.49 & 76.00 & 77.41 & 70.33 & 65.18 & 52.0 & 65.06 & 52.66 \\
\rowcolor{gray!10}
        Retrain & 95.67 & 0.00 & 97.22 & 0.00 & 77.18 & 0.00 & 78.01 & 0.00 & 66.36 & 0.00 & 66.73 & 0.00 \\
        FT & 93.93 & 0.00 & 93.68 & 0.00 & 71.77 & 3.00 & 71.89 & 0.00 & 62.61 & 2.00 & 59.82 & 3.33 \\
        NegGrad+ & 92.17 & 8.10 & 95.62 & 0.00 & 74.51 & 0.00 & 76.87 & 0.00 & 63.68 & 0.00 & 63.82 & 0.00 \\
        Random-label & 95.57 & 0.00 & 96.11 & 0.00 & 75.91 & 0.00 & 76.50 & 1.33 & 65.32 & 0.00 & 65.18 & 2.00 \\
        SalUn & 95.54 & 0.00 & 96.35 & 0.00 & 75.33 & 0.00 & 73.96 & 0.33 & 65.51 & 0.00 & 64.94 & 1.33 \\
        SCRUB & 95.66 & 0.00 & 96.58 & 0.00 & 74.58 & 0.00 & 76.54 & 0.00 & 63.12 & 0.00 & 63.10 & 0.00 \\
        UNSIR & 93.27 & 2.90 & 91.34 & 13.50 & 73.61 & 5.00 & 74.07 & 0.66 & 63.61 & 0.00 & 63.27 & 1.33 \\
        SSD & 85.24 & 38.76 & 76.18 & 28.26 & 74.28 & 0.00 & 75.21 & 0.00 & 64.06 & 0.00 & 63.38 & 0.00 \\
        \textbf{BiasShift} & 96.14 & 0.00 & \textbf{97.21} & 0.00 & 76.56 & 0.00 & 77.73 & 0.00 & \textbf{65.26} & 0.00 & \textbf{65.42} & 0.00 \\
        SF & 95.98 & 0.00 & 95.98 & 0.00 & 75.16 & 1.00 & 76.11 & 0.00 & 63.71 & 0.00 & 63.85 & 2.00 \\
        TS-BGM & 95.96 & 0.00 & 96.65 & 0.00 & \textbf{76.67} & 0.00 & 77.78 & 0.00 & 64.90 & 0.00 & 64.57 & 0.00 \\
        \textbf{TS-BGRM} & 96.\textbf{70} & 0.00 & 96.70 & 0.00 & \textbf{76.67} & 0.00 & 77.78 & 0.00 & 64.90 & 0.00 & 64.57 & 0.00 \\
        \textbf{LB-HR} & 95.89 & 0.00 & \textbf{97.11} & 0.00 & 76.45 & 0.00 & \textbf{77.81} & 0.00 & \textbf{65.37} & 0.00 & \textbf{65.32} & 0.00 \\
        \bottomrule
    \end{tabular}
\end{table*}

\begin{table*}[!htbp]
    \centering
    \caption{Model unlearning time and Recovery Time Ratio (RTR) results.}
    \label{tab:time}
    \begin{tabular}{@{}lccccccccccccc@{}}
        \toprule
        \multirow{3}{*}{\makecell{Method}} 
        & \multicolumn{4}{c}{CIFAR10} 
        & \multicolumn{4}{c}{CIFAR100} 
        & \multicolumn{4}{c}{Tiny-ImageNet} \\
           \cmidrule(lr){2-5} \cmidrule(lr){6-9} \cmidrule(lr){10-13}
        
        & \multicolumn{2}{c}{1} & \multicolumn{2}{c}{3} & \multicolumn{2}{c}{1} & \multicolumn{2}{c}{3} & \multicolumn{2}{c}{1} & \multicolumn{2}{c}{3} \\
         \cmidrule(lr){2-3} \cmidrule(lr){4-5}  \cmidrule(lr){6-7} \cmidrule(lr){8-9} \cmidrule(lr){10-11} \cmidrule(lr){12-13}
        & Time & RTR & Time & RTR & Time & RTR & Time & RTR & Time & RTR & Time & RTR \\
        \midrule
\rowcolor{gray!10}
        Retrain & 1972.01 & - & 1890.25 & - & 2088.29 & - & 1962.21 & - & 15776.07 & - & 15699.39 & - \\
        FT & 204.40 & 10.36 & 50.77 & 2.68 & 156.66 & 7.50 & 156.87 & 7.99 & 255.04 & 1.61 & 249.54 & 1.58 \\
        NegGrad+ & 189.41 & 9.60 & 85.45 & 4.52 & 210.89 & 10.09 & 166.93 & 8.50 & 363.42 & 2.30 & 555.83 & 3.54 \\
        Random-label & 91.92 & 4.66 & 68.05 & 3.60 & 42.61 & 2.04 & 65.44 & 3.33 & 162.13 & 1.02 & 164.88 & 1.05 \\
        SalUn & 47.26 & 2.39 & 49.37 & 2.61 & 48.85 & 2.33 & 49.26 & 2.51 & 164.53 & 1.04 & 141.76 & 0.90 \\
        SCRUB & 314.26 & 15.93 & 267.79 & 14.16 & 283.85 & 13.59 & 682.31 & 34.77 & 1216.76 & 7.71 & 1208.04 & 7.69 \\
        UNSIR & 76.93 & 3.90 & 82.09 & 4.34 & 136.40 & 6.53 & 141.87 & 7.23 & 718.40 & 4.55 & 828.97 & 5.28 \\
        SSD & 28.93 & 1.46 & 30.16 & 1.59 & 20.05 & 0.96 & 19.04 & 0.97 & 43.12 & 0.27 & 55.05 & 0.35 \\
 \multirow{2}{*}{\textbf{BiasShift}} 
& \multirow{2}{*}{\makecell{0.018}} 
& \multirow{2}{*}{\makecell{$9.12$ \\ $\times10^{-4}$}} 
& \multirow{2}{*}{\makecell{0.016}} 
& \multirow{2}{*}{\makecell{$8.46$ \\ $\times10^{-4}$}} 
& \multirow{2}{*}{\makecell{0.020}} 
& \multirow{2}{*}{\makecell{$9.57$ \\ $\times10^{-4}$}} 
& \multirow{2}{*}{\makecell{0.016}} 
& \multirow{2}{*}{\makecell{$8.15$ \\ $\times10^{-4}$}} 
& \multirow{2}{*}{\makecell{0.027}} 
& \multirow{2}{*}{\makecell{$1.71$ \\ $\times10^{-4}$}} 
& \multirow{2}{*}{\makecell{0.030}} 
& \multirow{2}{*}{\makecell{$1.91$ \\ $\times10^{-4}$}} \\
    \addlinespace
    \addlinespace
        SF & 69.40 & 3.51 & 50.40 & 2.66 & 170.25 & 8.15 & 169.78 & 8.65 & 298.46 & 1.89 & 294.98 & 1.87 \\
        TS-BGM & 10.17 & 0.51 & 9.51 & 0.50 & 11.44 & 0.54 & 11.55 & 0.58 & 33.85 & 0.21 & 32.18 & 0.20 \\
\rowcolor{orange!10}
        \textbf{TS-BGRM} & 11.37 & 0.57 & 9.57 & 0.50 & 11.85 & 0.56 & 11.51 & 0.58 & 32.86 & 0.20 & 32.86 & 0.20 \\
\rowcolor{orange!10}
        \textbf{LB-HR} & 40.14 & 2.03 & 37.22 & 1.96 & 91.15 & 4.36 & 83.61 & 4.26 & 152.47 & 0.96 & 153.67 & 0.97 \\
        \bottomrule
    \end{tabular}
\end{table*}

\begin{table}[!htbp]
  \centering
  \caption{Results of models on the BSC metric.}
  \label{tab:bsc}
  \begin{tabular}{@{}lcccccc@{}}
    \toprule
    Method & \multicolumn{2}{c}{CIFAR10} & \multicolumn{2}{c}{CIFAR100} & \multicolumn{2}{c}{Tiny-ImageNet} \\
    \cmidrule(lr){2-3} \cmidrule(lr){4-5} \cmidrule(lr){6-7}
           & 1     & 3     & 1     & 3     & 1     & 3     \\
    \midrule
\rowcolor{gray!10}
    Original  & 97.09 & 99.43 & 99.71 & 99.87 & 99.52 & 99.56 \\
\rowcolor{gray!10}
    Retrain   & 86.29 & 84.15 & 97.62 & 97.50 & 98.33 & 98.30 \\
    FT        & 93.29 & 91.13 & 97.51 & 97.01 & 98.97 & 98.53 \\
    NegGrad+  & 87.45 & 85.78 & 87.68 & 95.65 & 96.63 & 98.09 \\
    Random-label & 95.41 & 89.10 & 98.09 & 97.84 & 99.75 & 99.91 \\
    SalUn     & 94.48 & 85.49 & 96.65 & 96.54 & 99.71 & 99.96 \\
    SCRUB     & 61.08 & 61.54 & 85.26 & 95.39 & 83.64 & 93.88 \\
\rowcolor{blue!10}
    UNSIR     & 96.76 & 98.72 & 99.46 & 98.98 & 93.68 & 98.45 \\
\rowcolor{blue!10}
    SSD       & 99.50 & 99.66 & 99.98 & 99.99 & 99.99 & 99.98 \\
    BiasShift & 6.26  & 6.25  & 3.84  & 4.76  & 3.84  & 3.84  \\
    SF        & 88.91 & 86.29 & 91.95 & 89.68 & 92.43 & 92.80 \\
    TS-BGM    & 86.39 & 77.33 & 89.64 & 89.75 & 92.82 & 92.15 \\
\rowcolor{orange!10}
    TS-BGRM   & 97.22 & 99.56 & 99.32 & 99.86 & 99.66 & 99.78 \\
\rowcolor{orange!10}
    LB-HR     & 97.01 & 92.62 & 98.96 & 98.84 & 98.98 & 98.95 \\
    \bottomrule
  \end{tabular}
\end{table}

\begin{table}[!htbp]
  \centering
  \caption{Results of models on the MBG metric.}
  \label{tab:mbg}
  \setlength{\tabcolsep}{3pt}
  \begin{tabular}{@{}lcccccc@{}}
    \toprule
    Method & \multicolumn{2}{c}{CIFAR10} & \multicolumn{2}{c}{CIFAR100} & \multicolumn{2}{c}{Tiny-ImageNet} \\
    \cmidrule(lr){2-3} \cmidrule(lr){4-5} \cmidrule(lr){6-7}
           & 1     & 3     & 1     & 3     & 1     & 3     \\
    \midrule
\rowcolor{gray!10}
    Original  & 51.08 & 50.37 & 50.81 & 50.74 & 50.63 & 50.60 \\
\rowcolor{gray!10}
    Retrain   & 46.51 & 45.58 & 49.76 & 49.83 & 50.07 & 50.12 \\
    FT        & 48.52 & 47.76 & 50.16 & 49.98 & 50.28 & 50.11 \\
    NegGrad+  & 46.81 & 46.24 & 47.35 & 49.58 & 49.65 & 50.00 \\
    Random-label & 49.07 & 47.19 & 50.44 & 50.14 & 50.58 & 50.50 \\
    SalUn     & 48.90 & 46.46 & 50.05 & 49.77 & 50.58 & 50.48 \\
    SCRUB     & 35.03 & 36.75 & 46.58 & 49.50 & 45.81 & 48.93 \\
\rowcolor{blue!10}
    UNSIR     & 51.16 & 50.58 & 51.13 & 51.07 & 52.27 & 51.05 \\
\rowcolor{blue!10}
    SSD       & 50.46 & 50.34 & 50.88 & 50.61 & 50.51 & 50.50 \\
    \multirow{2}{*}{BiasShift}
    & \multirow{2}{*}{\makecell{$3.19$ \\ $\times10^{-4}$}}
    & \multirow{2}{*}{\makecell{$3.10$ \\ $\times10^{-4}$}}
    & \multirow{2}{*}{\makecell{$1.43$ \\ $\times10^{-8}$}}
    & \multirow{2}{*}{\makecell{$2.12$ \\ $\times10^{-6}$}}
    & \multirow{2}{*}{\makecell{$1.42$ \\ $\times10^{-8}$}}
    & \multirow{2}{*}{\makecell{$1.42$ \\ $\times10^{-8}$}} \\
    \addlinespace
    \addlinespace
    SF        & 47.45 & 46.46 & 49.59 & 48.81 & 49.95 & 50.08 \\
    TS-BGM    & 46.39 & 43.19 & 48.04 & 47.78 & 48.74 & 48.50 \\
\rowcolor{orange!10}
    TS-BGRM   & 51.05 & 50.59 & 50.95 & 50.76 & 50.91 & 50.89 \\
\rowcolor{orange!10}
    LB-HR     & 49.68 & 48.96 & 50.61 & 50.28 & 50.43 & 50.37 \\
    \bottomrule
  \end{tabular}
\end{table}

\begin{table}[!htbp]
  \centering
  \caption{Results of models on the MBS metric.}
  \label{tab:mbs} 
  \setlength{\tabcolsep}{3pt}
  \begin{tabular}{@{}lcccccc@{}}
    \toprule
    Method & \multicolumn{2}{c}{CIFAR10} & \multicolumn{2}{c}{CIFAR100} & \multicolumn{2}{c}{Tiny-ImageNet} \\
    \cmidrule(lr){2-3} \cmidrule(lr){4-5} \cmidrule(lr){6-7}
           & 1     & 3     & 1     & 3     & 1     & 3     \\
    \midrule
\rowcolor{gray!10}
    Original  & 51.08 & 49.91 & 50.81 & 50.42 & 50.63 & 50.34 \\
\rowcolor{gray!10}
    Retrain   & 46.51 & 45.58 & 49.76 & 49.83 & 50.07 & 50.12 \\
    FT        & 48.52 & 47.49 & 50.16 & 49.70 & 50.28 & 49.87 \\
    NegGrad+  & 46.81 & 45.73 & 47.35 & 49.27 & 49.65 & 49.74 \\
    Random-label & 49.07 & 46.91 & 50.44 & 49.81 & 50.58 & 50.25 \\
    SalUn     & 48.90 & 46.15 & 50.05 & 49.45 & 50.58 & 50.23 \\
    SCRUB     & 35.03 & 36.53 & 46.58 & 49.24 & 45.81 & 48.83 \\
\rowcolor{blue!10}
    UNSIR     & 51.16 & 50.15 & 51.13 & 50.74 & 52.27 & 50.85 \\
\rowcolor{blue!10}
    SSD       & 50.46 & 50.19 & 50.88 & 50.61 & 50.51 & 50.49 \\
    \multirow{2}{*}{BiasShift}
& \multirow{2}{*}{\makecell{$3.19$ \\ $\times10^{-4}$}}
& \multirow{2}{*}{\makecell{$3.04$ \\ $\times10^{-4}$}}
& \multirow{2}{*}{\makecell{$1.43$ \\ $\times10^{-8}$}}
& \multirow{2}{*}{\makecell{$2.09$ \\ $\times10^{-6}$}}
& \multirow{2}{*}{\makecell{$1.42$ \\ $\times10^{-8}$}}
& \multirow{2}{*}{\makecell{$1.40$ \\ $\times10^{-8}$}} \\
\addlinespace
\addlinespace
    SF        & 47.45 & 45.81 & 49.59 & 48.68 & 49.95 & 49.98 \\
    TS-BGM    & 46.39 & 42.65 & 48.04 & 47.47 & 48.74 & 48.25 \\
\rowcolor{orange!10}
    TS-BGRM   & 51.05 & 50.04 & 50.95 & 50.44 & 50.91 & 50.64 \\
\rowcolor{orange!10}
    LB-HR     & 49.68 & 48.96 & 50.61 & 50.28 & 50.43 & 50.37 \\
    \bottomrule
  \end{tabular}
\end{table}

\begin{figure*}[!t]
	\centering 
	\subfigure[3rd--5th classes: bias values.]{
	\includegraphics[width=0.26\linewidth]{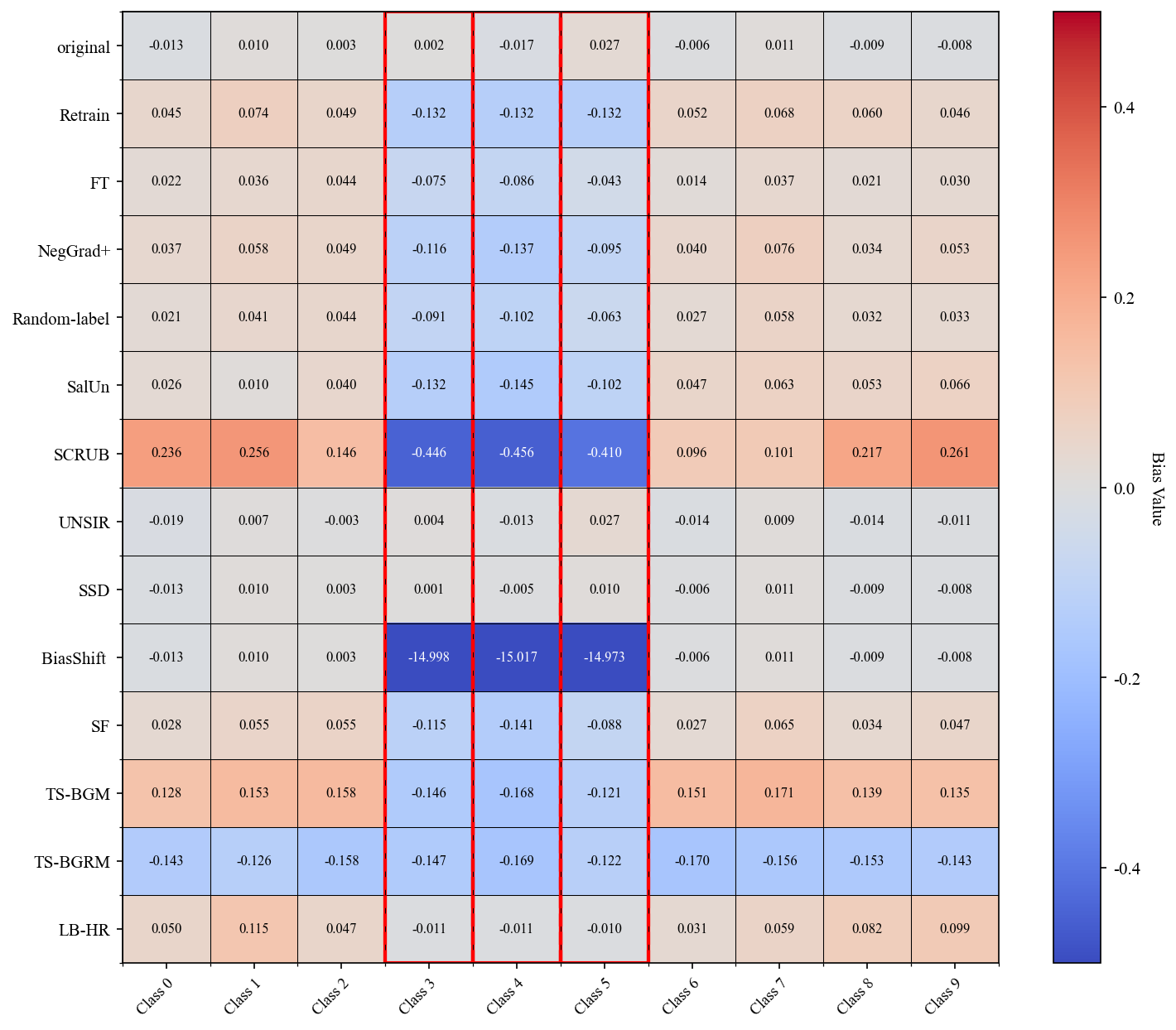}
	\label{fig:8a}
	}
	\subfigure[3rd--5th classes: difference.]{
	\includegraphics[width=0.21\linewidth]{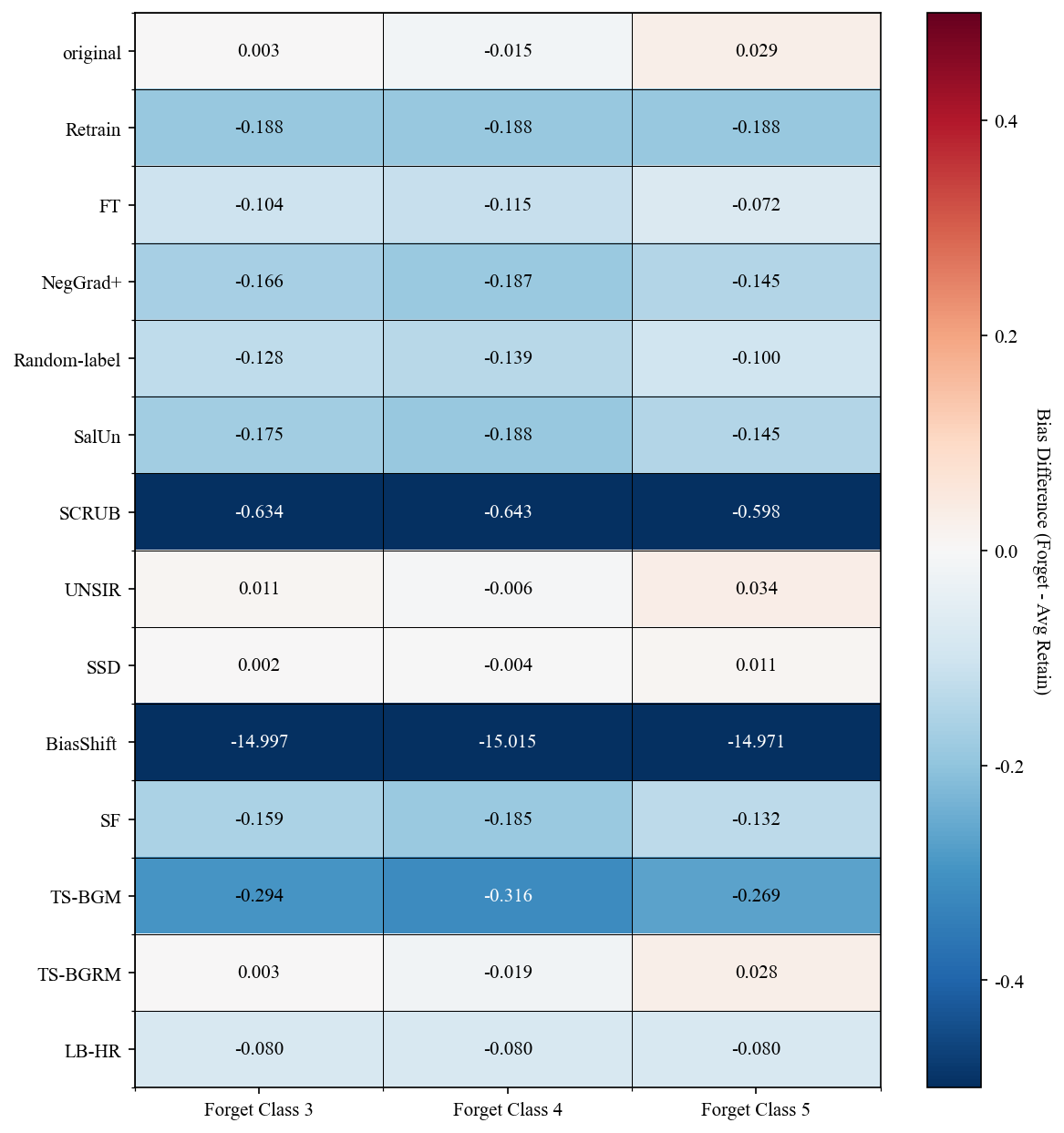}
	\label{fig:8b}
	}
	\subfigure[5th class: bias value.]{
	\includegraphics[width=0.26\linewidth]{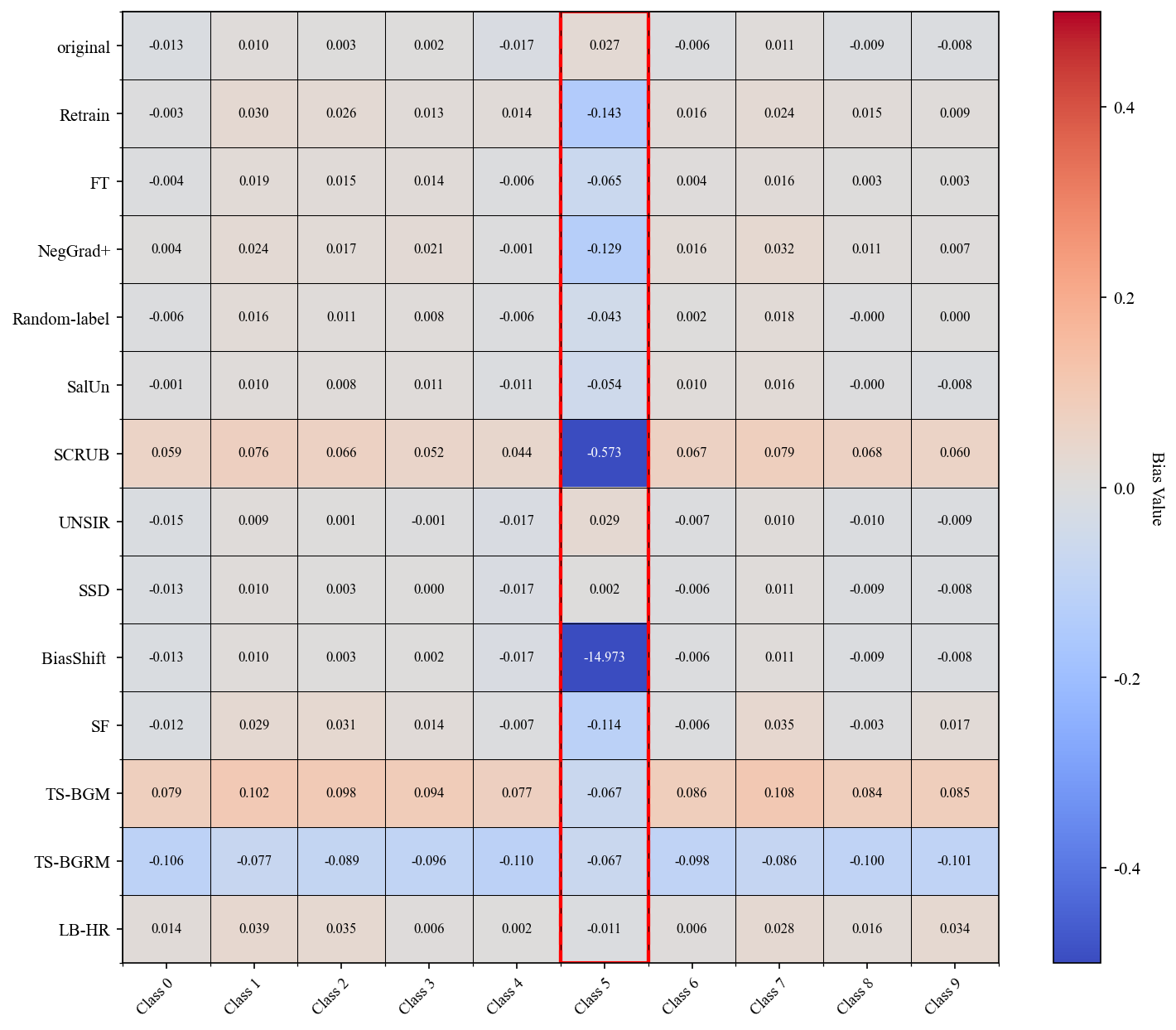}
	\label{fig:8c}
	}
	\subfigure[5th class: difference.]{
	\includegraphics[width=0.21\linewidth]{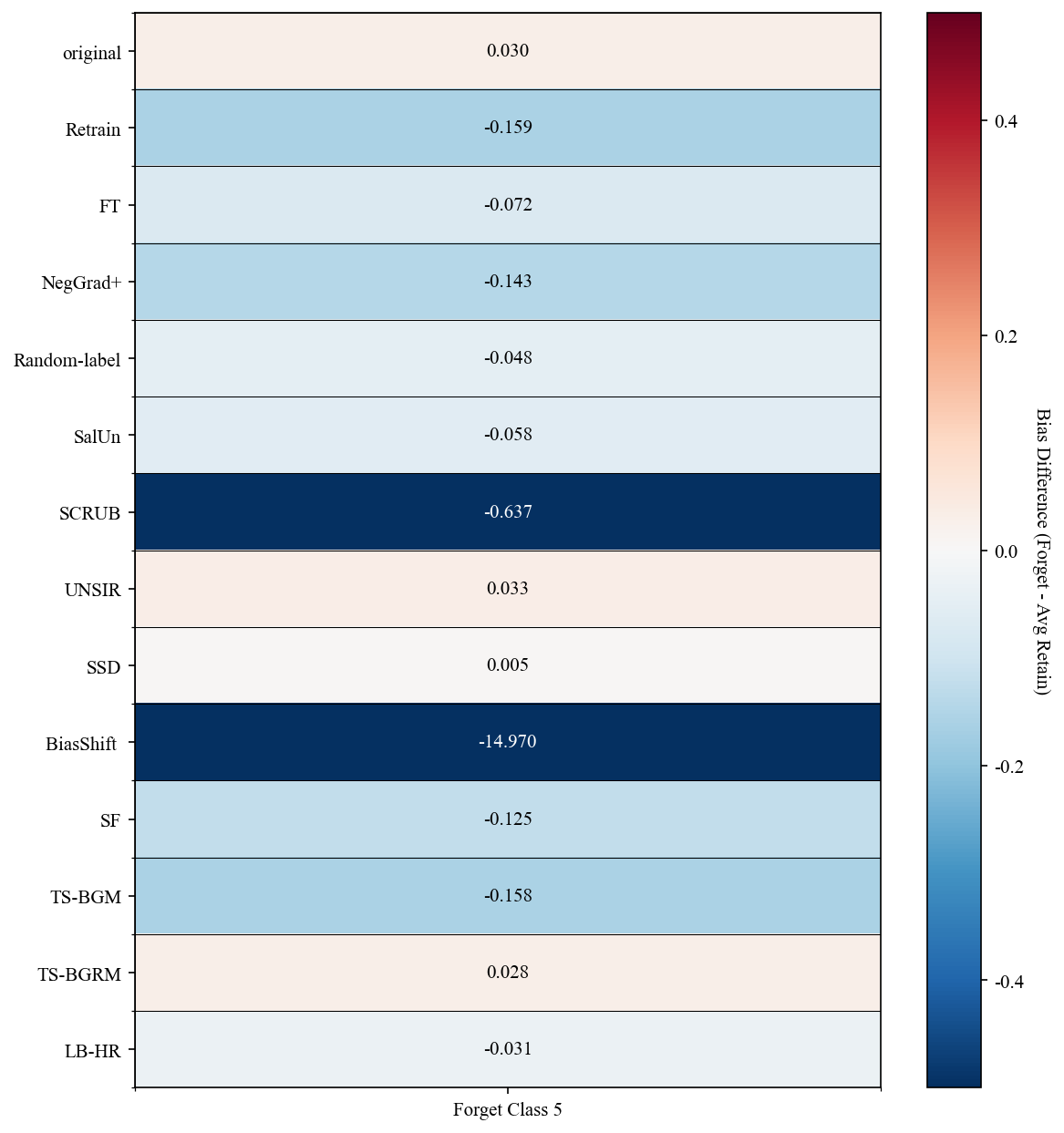}
	\label{fig:8d}
	}	
\caption{Classification-head bias analysis on CIFAR-10 after class-level forgetting. 
Subfigures (a)--(b) show the three-class forgetting setting, and subfigures (c)--(d) show the single-class forgetting setting.}
	\label{fig:8}
\end{figure*}

\begin{figure*}[!t]
	\centering 
	\subfigure[3rd--5th classes: bias values.]{
	\includegraphics[width=0.26\linewidth]{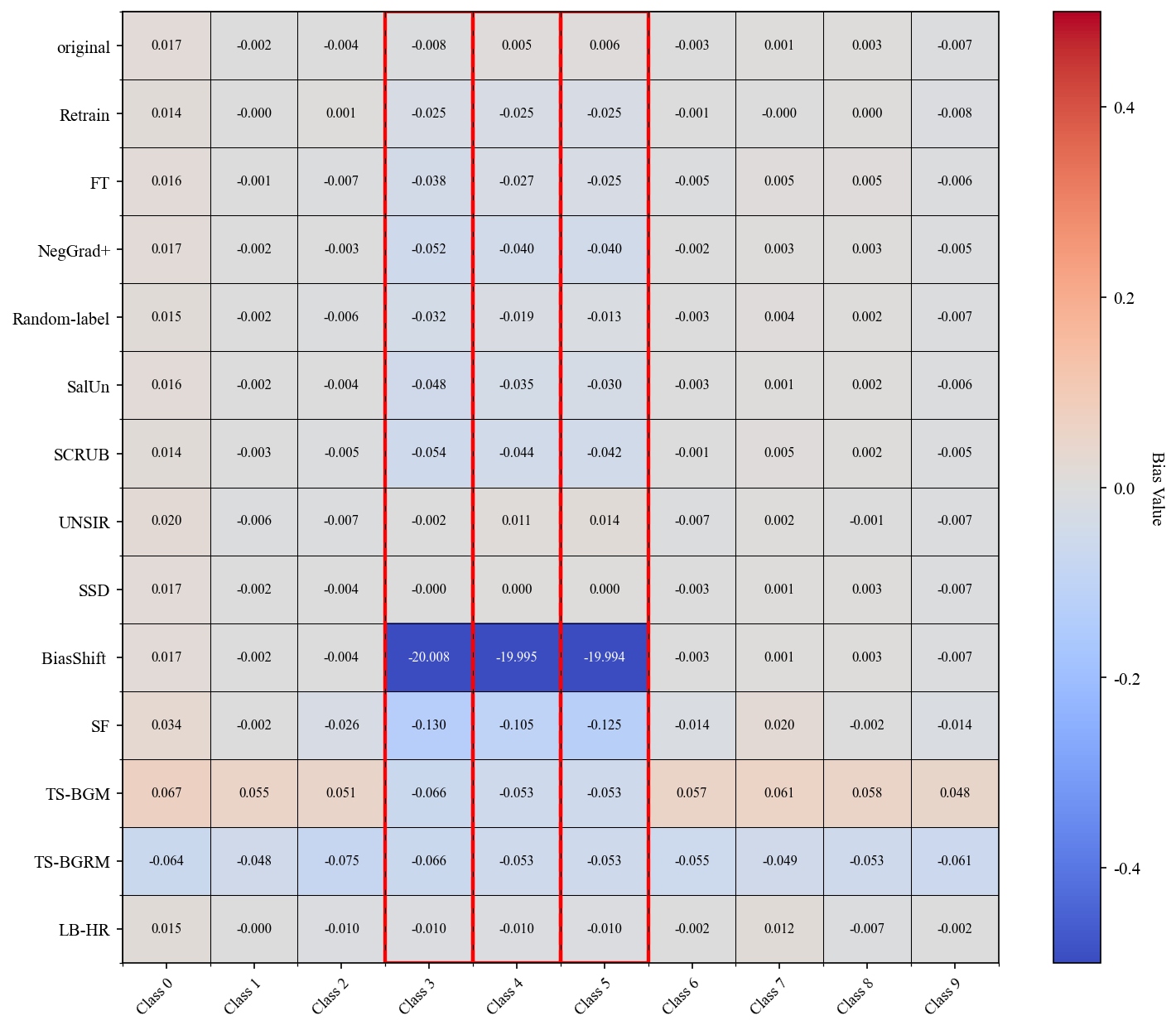}
	\label{fig:10a}
	}
	\subfigure[3rd--5th classes: difference.]{
	\includegraphics[width=0.21\linewidth]{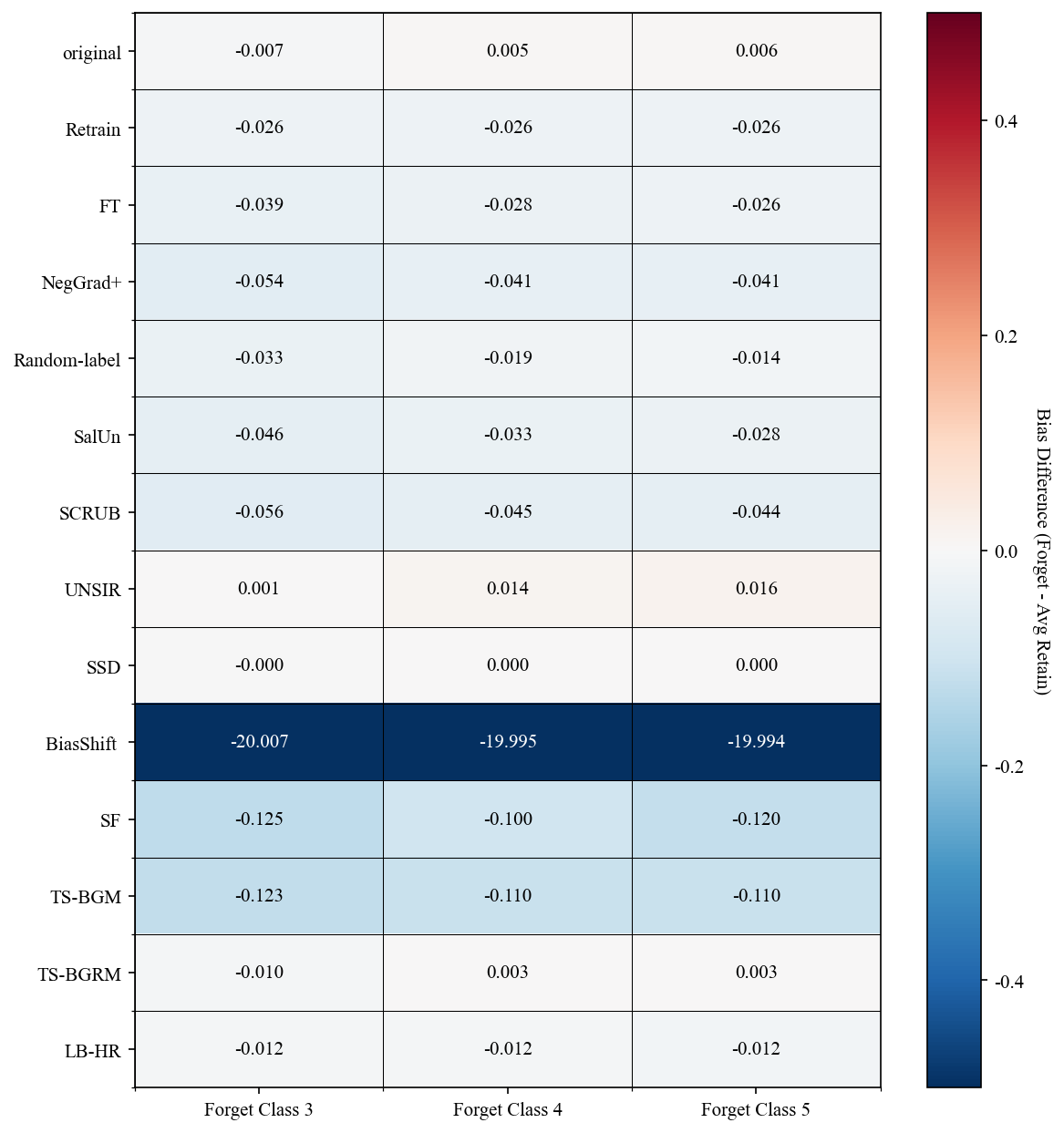}
	\label{fig:10b}
	}
	\subfigure[5th class: bias value.]{
	\includegraphics[width=0.26\linewidth]{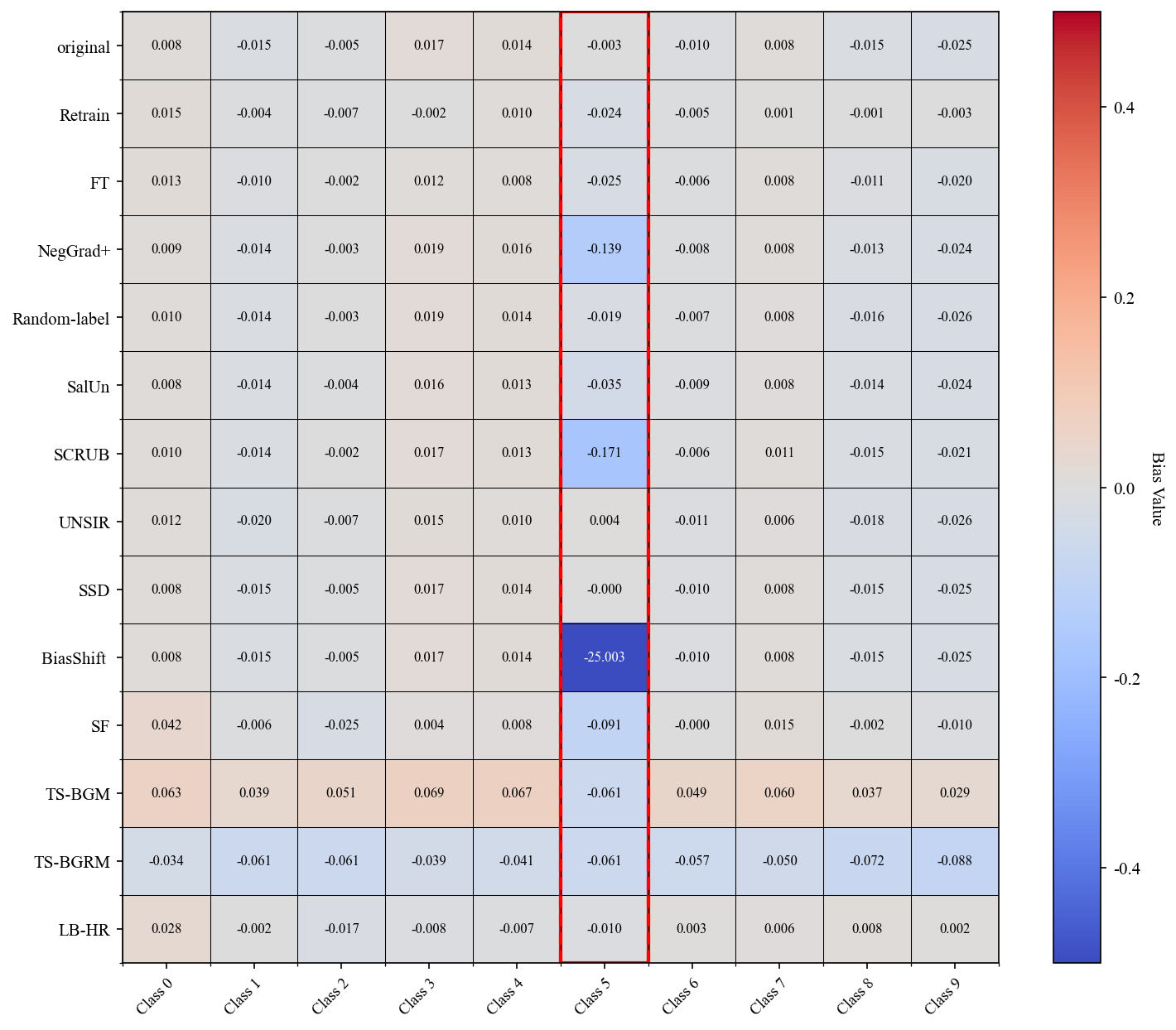}
	\label{fig:10c}
	}
	\subfigure[5th class: difference.]{
	\includegraphics[width=0.21\linewidth]{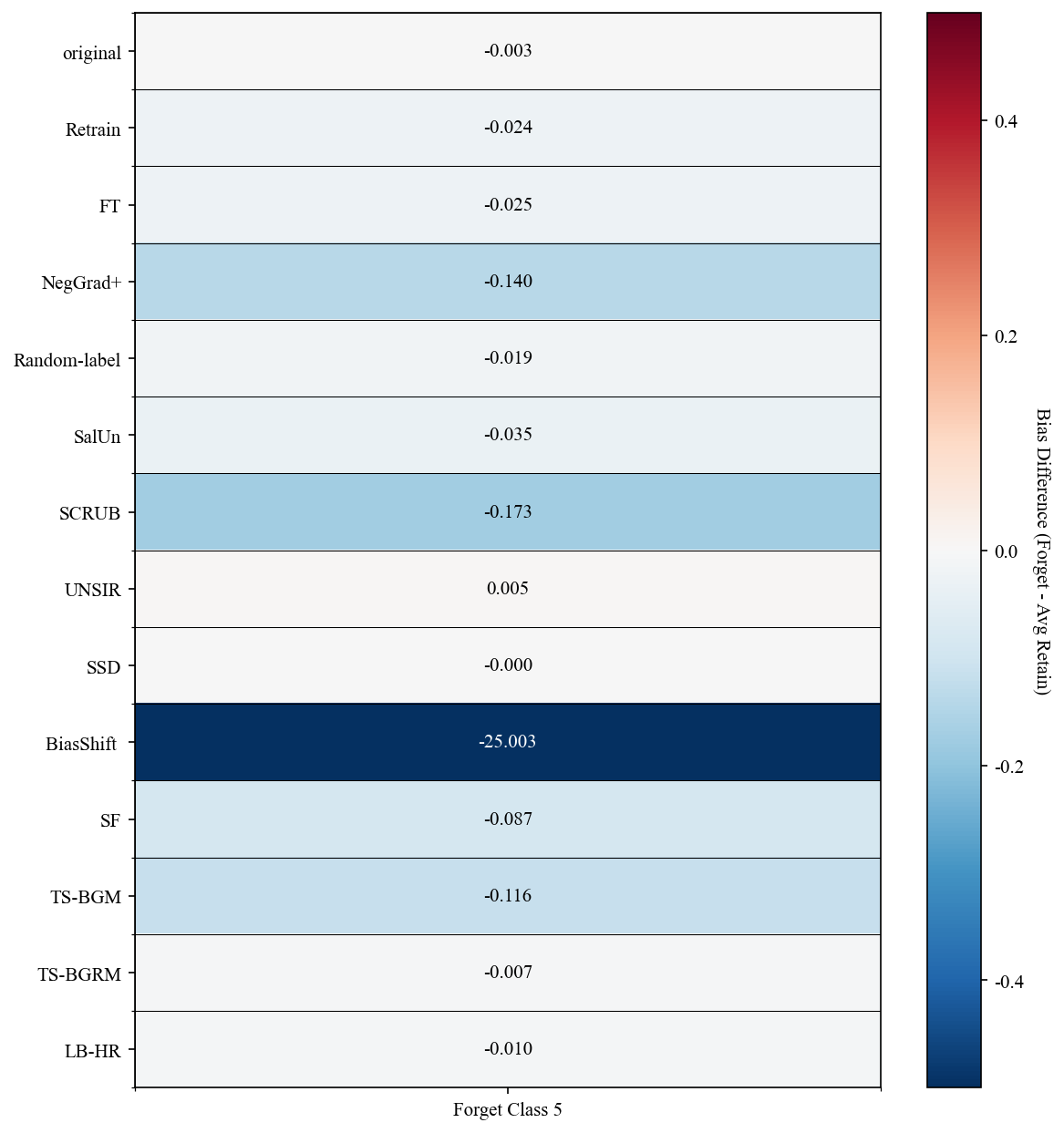}
	\label{fig:10d}
	}	
\caption{Classification-head bias analysis on CIFAR-100 after class-level forgetting. 
For clarity, only the first 10 class heads are visualized.}
	\label{fig:10}
\end{figure*}

\begin{figure*}[!t]
	\centering 
	\subfigure[3rd--5th classes: bias values.]{
	\includegraphics[width=0.26\linewidth]{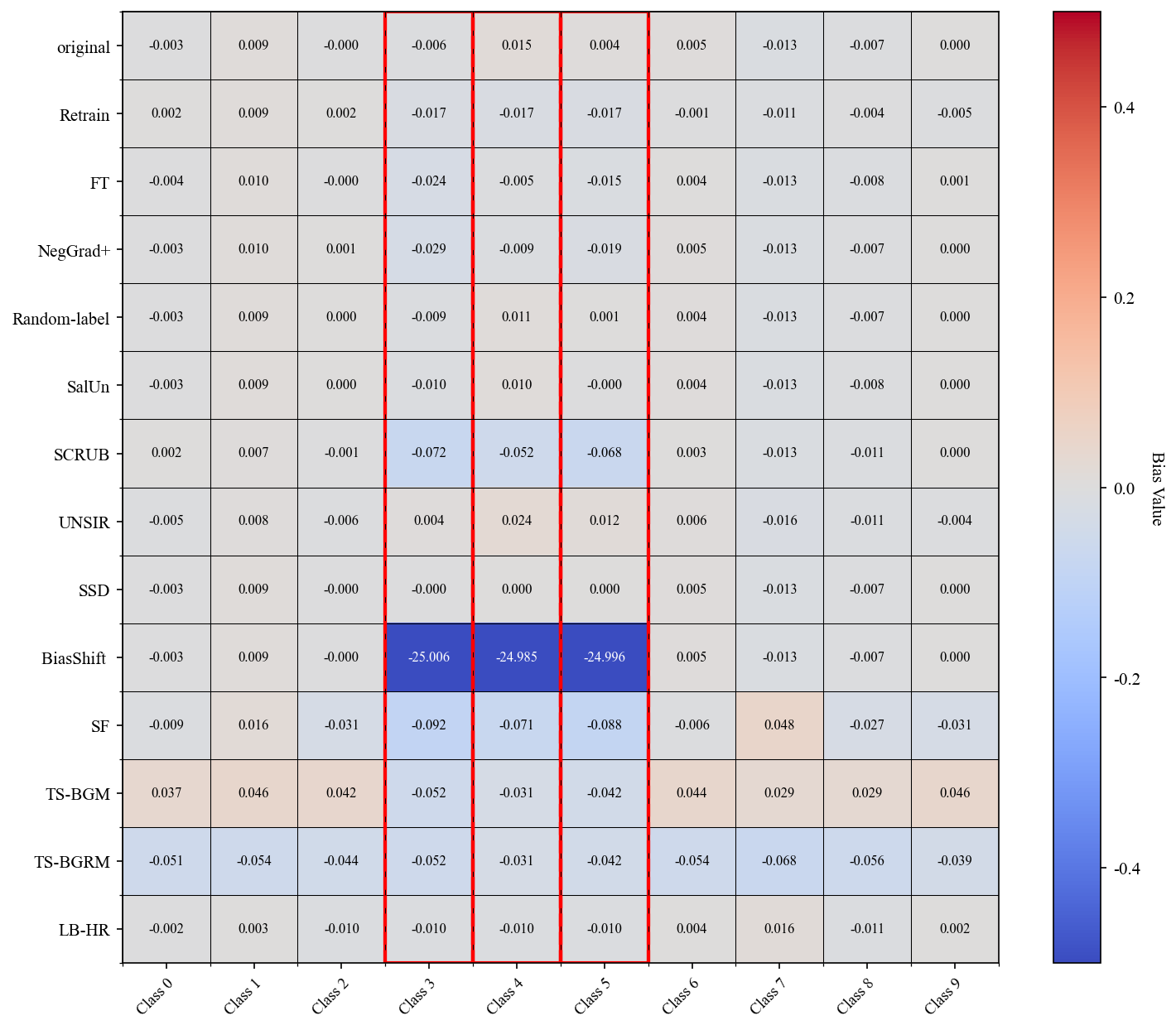}
	\label{fig:12a}
	}
	\subfigure[3rd--5th classes: difference.]{
	\includegraphics[width=0.21\linewidth]{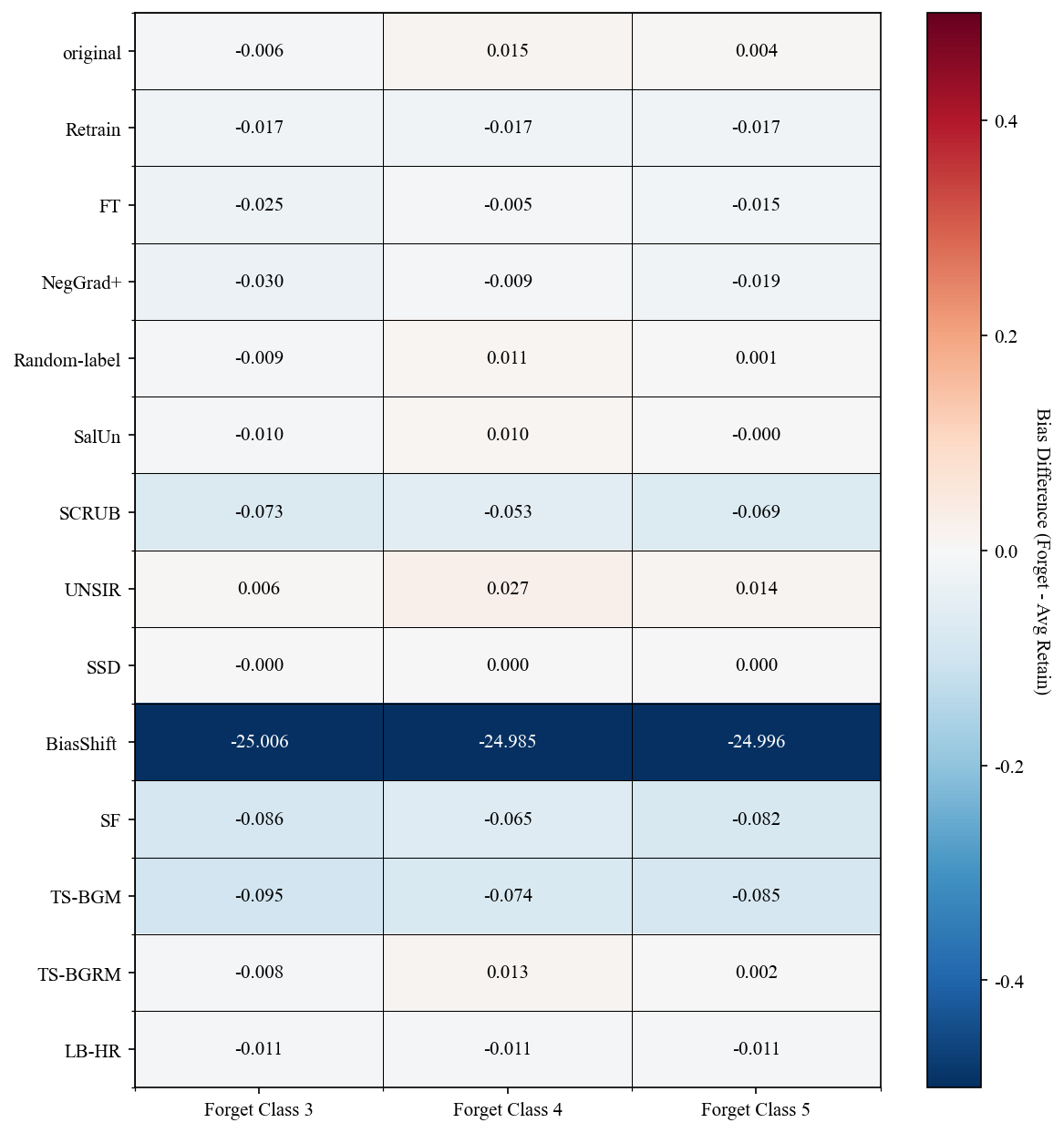}
	\label{fig:12b}
	}
	\subfigure[5th class: bias value.]{
	\includegraphics[width=0.26\linewidth]{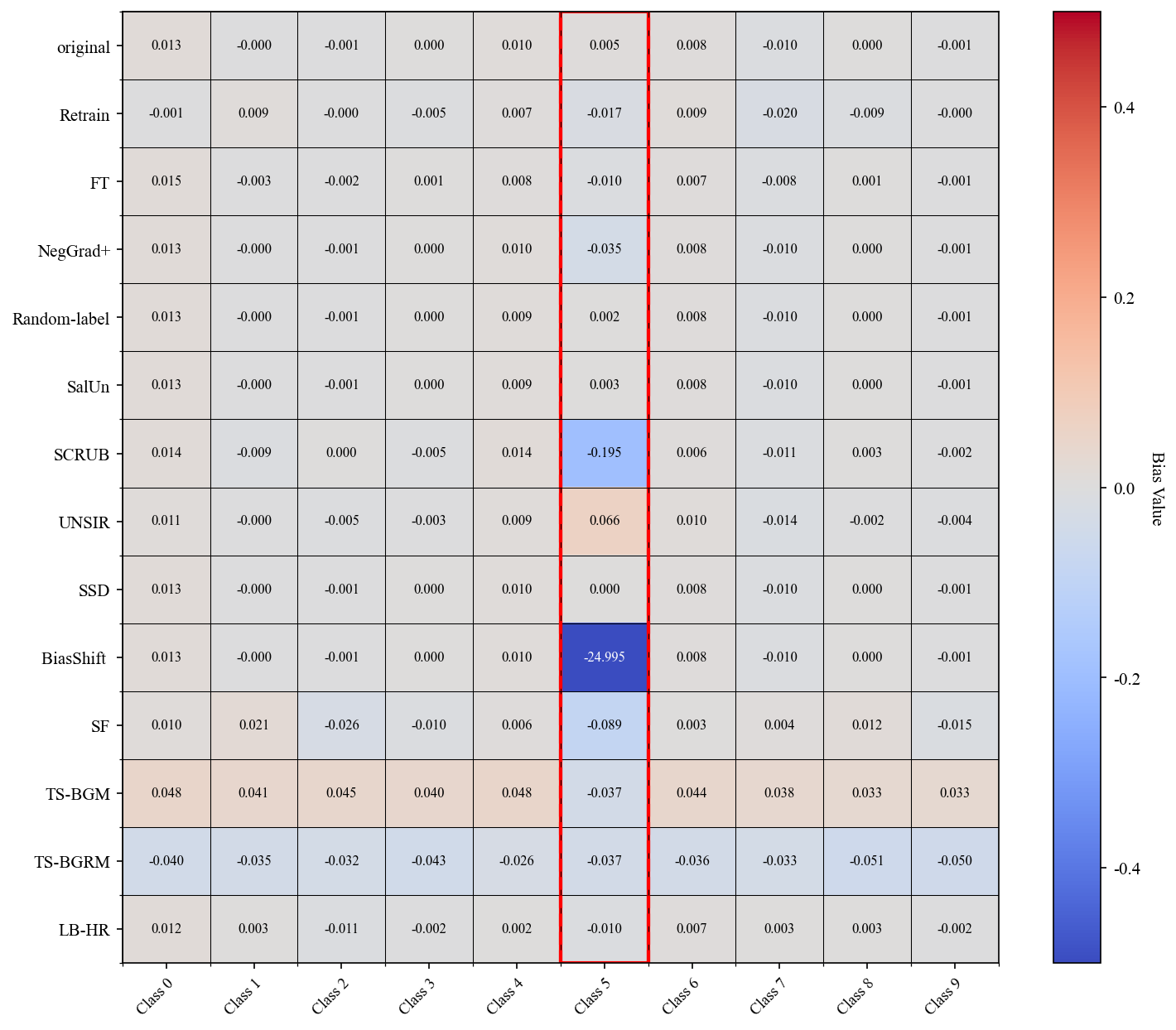}
	\label{fig:12c}
	}
	\subfigure[5th class: difference.]{
	\includegraphics[width=0.21\linewidth]{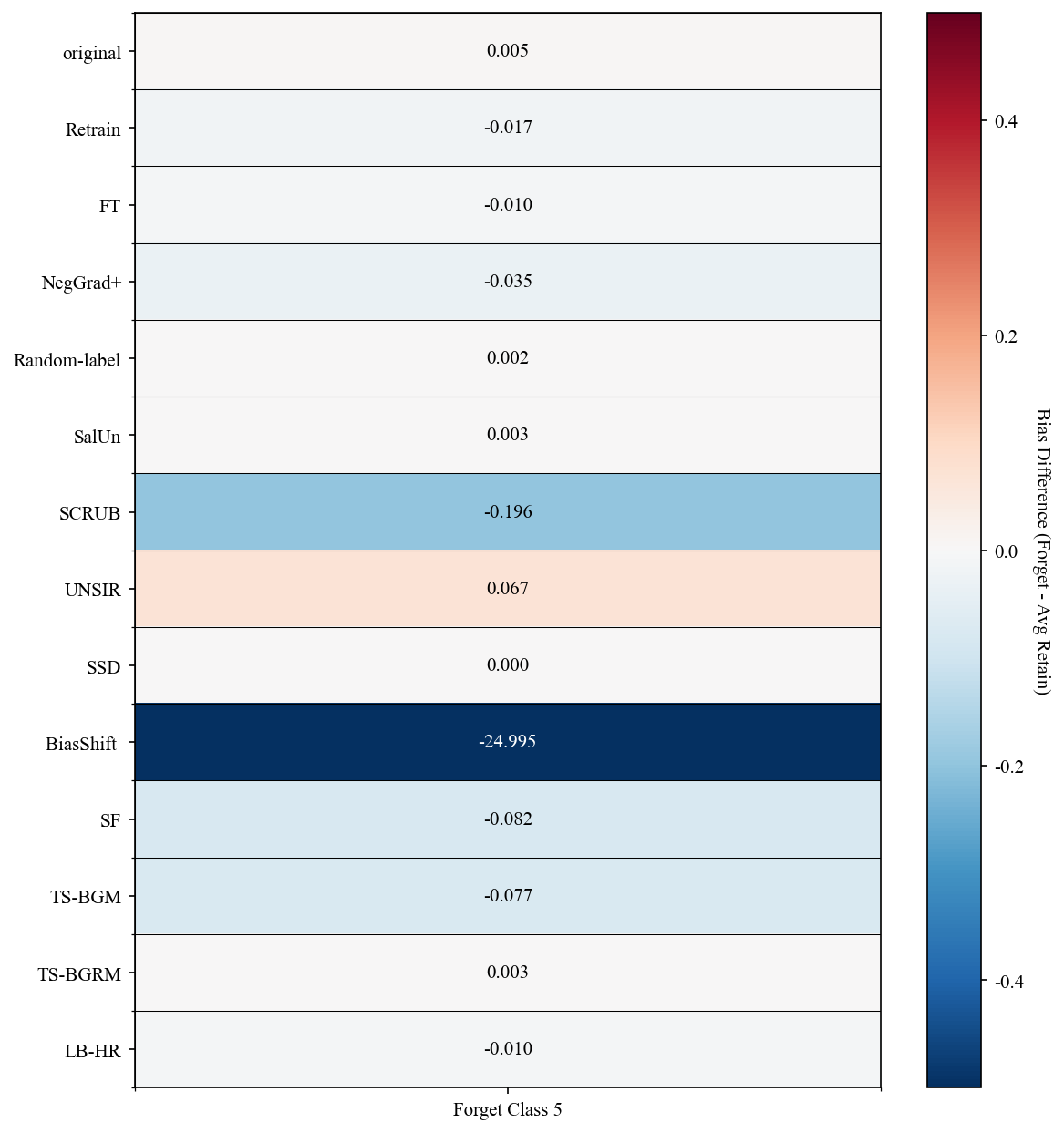}
	\label{fig:12d}
	}	
\caption{Classification-head bias analysis on Tiny-ImageNet after class-level forgetting. 
For clarity, only the first 10 class heads are visualized.}
	\label{fig:12}
\end{figure*}

\subsection{Experimental Results}
\label{subsec:experimental_results}

This subsection reports the experimental results from two complementary perspectives. 
We first analyze the conventional unlearning performance, including retain-set accuracy, forget-set accuracy, and unlearning time. 
We then examine the bias-oriented metrics and bias distributions to evaluate whether different methods rely on abnormal forgotten-class bias suppression.

\subsubsection{Conventional Unlearning Performance}

Table~\ref{tab:accuracy} reports the retain-set and forget-set accuracies of different methods on CIFAR-10, CIFAR-100, and Tiny-ImageNet under both single-class and three-class forgetting settings. 
The results show that many existing methods can reduce the forget-set accuracy to zero or near zero while maintaining reasonable retain-set accuracy. 
This indicates that class-level forgetting can often be achieved under conventional metrics. 
However, these metrics alone do not explain how the forgetting effect is realized internally.

A particularly important observation is that BiasShift achieves strong performance under conventional metrics. 
By only shifting the forgotten-class biases, BiasShift reduces the forget-set accuracy to zero in all evaluated settings while preserving competitive retain-set accuracy. 
For example, on CIFAR-10, BiasShift achieves $96.14\%$ retain-set accuracy and $0.00\%$ forget-set accuracy in the single-class forgetting setting, and $97.21\%$ retain-set accuracy and $0.00\%$ forget-set accuracy in the three-class forgetting setting. 
Similar results can also be observed on CIFAR-100 and Tiny-ImageNet. 
These results verify RQ1: a simple bias-level manipulation can satisfy conventional class-level unlearning metrics.

Nevertheless, the strong performance of BiasShift should not be interpreted as evidence that it is a reliable or privacy-preserving unlearning method. 
Instead, it demonstrates that retain-set accuracy and forget-set accuracy can be satisfied through an output-layer shortcut. 
Since BiasShift does not modify the feature extractor or the classification weights, its success under conventional metrics suggests that these metrics may overestimate the reliability of class-level unlearning.

The proposed TS-BGRM and LB-HR methods also achieve competitive conventional unlearning performance. 
Across the three datasets, both methods maintain high retain-set accuracy and reduce the forget-set accuracy to zero in most settings. 
Compared with shallow fine-tuning (SF), TS-BGRM and LB-HR generally achieve better or comparable retain-set accuracy, indicating that their effectiveness is not merely due to freezing the feature extractor and updating fewer parameters. 
This supports the effectiveness of the proposed bias-aware mechanisms.

Table~\ref{tab:time} reports the unlearning time and Recovery Time Ratio (RTR). 
Retraining requires the largest time cost because it trains a new model from scratch on the retain set. 
Most approximate unlearning methods substantially reduce the time cost compared with retraining. 
BiasShift has the lowest time cost because it only modifies a few bias entries and does not involve iterative optimization. 
TS-BGRM is also highly efficient because it updates only the classification head in a short two-stage process. 
LB-HR requires more time than TS-BGRM but remains much faster than full retraining in most settings. 
These results show that the proposed methods preserve the efficiency advantage expected from approximate unlearning.

Overall, the conventional metrics in Tables~\ref{tab:accuracy} and~\ref{tab:time} show that BiasShift, TS-BGRM, and LB-HR can achieve competitive class-level unlearning performance. 
However, since BiasShift can also perform well under these metrics despite being a direct bias manipulation, further analysis is required to determine whether the forgetting effect depends on abnormal forgotten-class bias suppression.

\subsubsection{Bias-Oriented Evaluation}

Tables~\ref{tab:bsc}, \ref{tab:mbg}, and~\ref{tab:mbs} report the proposed bias-oriented metrics, including BSC, MBG, and MBS. 
These metrics reveal clear differences among methods that may appear similar under conventional accuracy metrics.

First, BiasShift obtains extremely low BSC, MBG, and MBS values across all datasets and forgetting settings. 
This is expected because BiasShift directly subtracts a large constant from the forgotten-class biases, making them much smaller than the retained-class biases. 
Although BiasShift achieves zero forget-set accuracy, its bias-oriented metrics clearly indicate severe forgotten-class bias suppression. 
This confirms that BiasShift leaves an obvious parameter-level signature of the forgotten labels.

Second, several existing unlearning methods also show different degrees of bias dependence. 
Methods based on retain-set optimization or gradient manipulation may reduce forget-set accuracy by suppressing the forgotten-class heads, resulting in lower BSC, MBG, or MBS values. 
For example, fine-tuning, NegGrad+, SalUn, SCRUB, and TS-BGM can achieve satisfactory forget-set accuracy in several settings, but their bias-oriented scores are lower than those of the proposed TS-BGRM and LB-HR in many cases. 
This suggests that low forget-set accuracy does not necessarily imply stable or non-revealing internal bias patterns. 
These results answer RQ2: existing class-level unlearning methods exhibit different degrees of dependence on forgotten-class bias suppression.

Third, TS-BGRM achieves consistently strong bias-oriented performance. 
Compared with its ablation variant TS-BGM, TS-BGRM obtains much higher BSC, MBG, and MBS values in most settings, although their retain-set and forget-set accuracies may be similar. 
This comparison is important because it shows the specific value of bias-gradient reversal. 
The reversal operation does not necessarily lead to a large improvement in conventional accuracy metrics, but it significantly improves the stability of the classification-head bias distribution. 
Therefore, the advantage of TS-BGRM is better captured by bias-oriented metrics than by retain-set or forget-set accuracy alone.

Fourth, LB-HR also improves bias stability by preventing forgotten-class biases from drifting toward extreme negative values. 
Its BSC values remain high across datasets, and its MBG and MBS values are generally close to or above the favorable threshold of $50\%$. 
This indicates that the forgotten-class biases are not significantly lower than the lower range of retained-class biases. 
Compared with TS-BGRM, LB-HR provides a more direct loss-level constraint on forgotten-class bias values, while TS-BGRM achieves bias stabilization through a two-stage optimization process. 
Both methods reduce abnormal forgotten-class bias suppression while maintaining competitive conventional unlearning performance, which answers RQ3.

\subsubsection{Visualization of Classification-Head Bias}

Figures~\ref{fig:8}, \ref{fig:10}, and~\ref{fig:12} visualize the classification-head bias distributions after unlearning on CIFAR-10, CIFAR-100, and Tiny-ImageNet, respectively. 
These visualizations provide intuitive evidence for the conclusions drawn from BSC, MBG, and MBS.
For each dataset, subfigures (a) and (b) correspond to the three-class forgetting setting, while subfigures (c) and (d) correspond to the single-class forgetting setting. 
The left subfigure in each pair shows the bias values, and the right subfigure shows the bias difference between forgotten-class heads and the average retained-class head.

For BiasShift, the forgotten-class biases are clearly separated from the retained-class biases because they are directly shifted toward large negative values. 
This explains why BiasShift obtains extremely low bias-oriented scores despite its strong conventional unlearning performance. 
The abnormal bias pattern makes the forgotten labels easily inferable from the final classification head.

For TS-BGM, the forgotten-class biases are still noticeably lower than the retained-class biases in several settings. 
This shows that simply using a two-stage procedure without bias-gradient reversal is insufficient to fully mitigate forgotten-class bias suppression. 
In contrast, TS-BGRM produces a more balanced bias distribution. 
The forgotten-class biases remain closer to the retained-class biases, reducing the risk that the forgotten labels can be identified as outliers.

For LB-HR, the forgotten-class biases are constrained around the lower-bound region rather than drifting toward extreme negative values. 
This confirms the effect of the lower-bound hinge regularization term. 
Although LB-HR and TS-BGRM use different mechanisms, both methods reduce the abnormal separation between forgotten-class and retained-class biases.

These visual results further support RQ4. 
Conventional metrics can show whether a model predicts forgotten classes incorrectly, but they cannot reveal whether this behavior is caused by an abnormal final-layer bias pattern. 
The proposed bias-oriented metrics and visualizations provide complementary internal evidence for evaluating the reliability of class-level unlearning.

\section{Conclusion}
\label{sec:conclusion}

In this paper, we investigated class-level machine unlearning from the perspective of classification-head bias. 
We showed that the bias terms of the final classification head provide an input-independent shortcut for suppressing class logits, and explained why the biases of absent classes tend to decrease during retain-set-only optimization. 
This analysis reveals that conventional class-level unlearning metrics may be satisfied through forgotten-class bias suppression rather than reliable removal of class-related influence.

Based on this observation, we introduced BiasShift as a diagnostic baseline to expose the bias-dominated shortcut in conventional unlearning evaluation. 
Although BiasShift can achieve high retain-set accuracy, near-zero forget-set accuracy, and negligible unlearning time, it also leaves abnormal bias patterns that make the forgotten labels inferable from the classification head. 
To mitigate this problem, we proposed TS-BGRM and LB-HR, which reduce excessive forgotten-class bias suppression through bias-gradient reversal and lower-bound hinge regularization, respectively. 
We further introduced BSC, MBG, and MBS to quantify bias dependence and potential forgotten-label leakage.

Experiments on CIFAR-10, CIFAR-100, and Tiny-ImageNet demonstrate that conventional metrics alone cannot distinguish reliable unlearning from bias-level output suppression. 
The proposed TS-BGRM and LB-HR maintain competitive unlearning performance while producing more stable and less revealing bias distributions. 
These results suggest that class-level machine unlearning should be evaluated from both external prediction behavior and internal parameter evidence. 
Future work will extend the proposed bias-aware analysis to broader architectures, tasks, and privacy evaluations.


\end{document}